\documentclass{article}
\PassOptionsToPackage{square, numbers}{natbib} 
\usepackage[preprint]{neurips_2024}

\usepackage[utf8]{inputenc} 
\usepackage[T1]{fontenc}    
\usepackage{hyperref}       
\usepackage{url}            
\usepackage{booktabs}       
\usepackage{amsfonts}       
\usepackage{nicefrac}       
\usepackage{microtype}      
\usepackage{xcolor}         
\usepackage[ruled,vlined]{algorithm2e}
\usepackage{graphicx}
\usepackage{soul}
\usepackage{tikz}
\usepackage{amssymb}

\usepackage{wrapfig}
\usepackage{caption}
\usepackage{mathtools}
\usepackage{multirow}
\usepackage{algpseudocode}
\usepackage{comment}
\usepackage{subcaption}
\usepackage{listings}
\usepackage{float}
\usepackage{enumitem}
\setlist[enumerate]{itemsep=0mm}
\setlist{nosep}

\newcommand{\mmct}{{\texttt{MMCTAgent}}\xspace}

\title{\mmct: Multi-modal Critical Thinking Agent Framework for Complex Visual Reasoning}

\newcommand*\samethanks[1][\value{footnote}]{\footnotemark[#1]}

\author{%
Somnath Kumar\thanks{Equal Contributions} \quad Yash Gadhia\samethanks \quad Tanuja Ganu\quad Akshay Nambi \\
Microsoft Research India \\
\texttt{\{akshayn, taganu\}@microsoft.com}\\
}

\begin{document}

\maketitle

\begin{abstract}

Recent advancements in Multi-modal Large Language Models (MLLMs) have significantly improved their performance in tasks combining vision and language. However, challenges persist in detailed multi-modal understanding, comprehension of complex tasks, and reasoning over multi-modal information. This paper introduces \mmct, a novel multi-modal critical thinking agent framework designed to address the inherent limitations of current MLLMs in complex visual reasoning tasks. Inspired by human cognitive processes and critical thinking, \mmct iteratively analyzes multi-modal information, decomposes queries, plans strategies, and dynamically evolves its reasoning. Additionally, \mmct incorporates critical thinking elements such as verification of final answers and self-reflection through a novel approach that defines a vision-based critic and identifies task-specific evaluation criteria, thereby enhancing its decision-making abilities. Through rigorous evaluations across various image and video understanding benchmarks, we demonstrate that \mmct (with and without the critic) outperforms both foundational MLLMs and other tool-augmented pipelines.


\end{abstract}

\section{Introduction}
\label{sec:intro}
\begin{figure*}[!t]
\begin{minipage}[t]{0.5\linewidth}
        \includegraphics[width=1.0\columnwidth,height=3in]{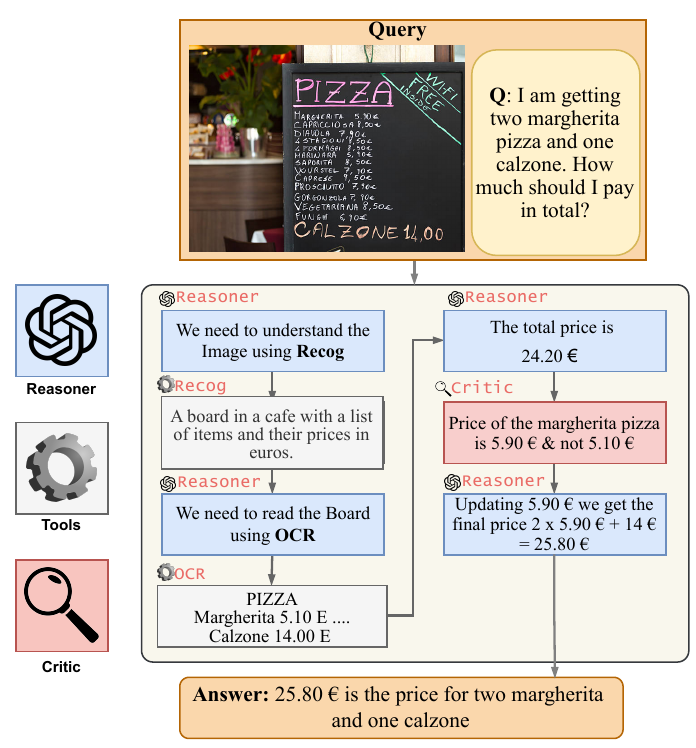}
        \caption{\mmct: Image understanding.}
        \label{fig:image_overview}
        \vspace{-10pt}
\end{minipage}
~
\begin{minipage}[t]{0.5\linewidth}
        \includegraphics[width=1.0\columnwidth,height=3in ]{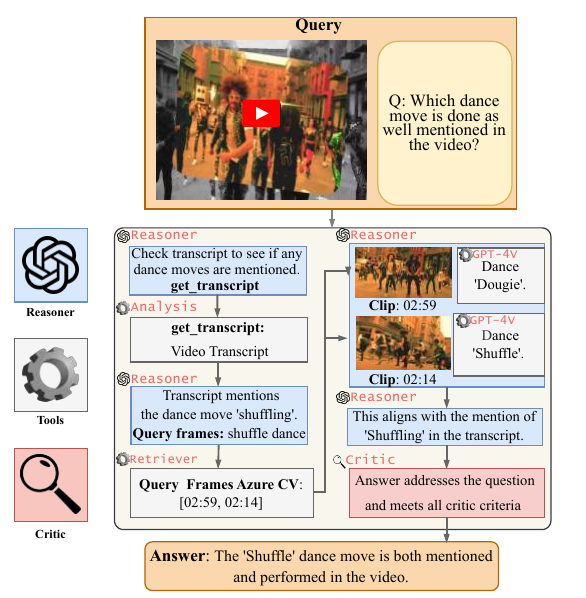}
        \caption{\mmct: Video understanding.}
        \label{fig:video_overview}
        \vspace{-10pt}
\end{minipage}
\vspace{-15pt}
\end{figure*}

Recent advancements in Multi-modal Large Language Models (MLLMs), such as GPT-4-Vision~\cite{achiam2023gpt4}, Gemini~\cite{team2023gemini}, and Qwen VL~\cite{bai2023qwen}, have significantly improved performance in vision and language tasks, allowing zero-shot problem-solving with images and videos~\cite{wang2023image}. One crucial task is Visual Question Answering (VQA)~\cite{agrawal2016vqa}, requiring comprehension and reasoning over multi-modal information to answer questions about images or long-form videos, spanning from minutes to hours.
Despite recent advancements, MLLMs still have inherent limitations in detailed multi-modal processing (e.g., spatial understanding, limited context length), comprehending complex tasks, and reasoning over multi-modal information, constraining their practical applicability~\cite{2023opencompass}. Figure~\ref{fig:image_overview} exemplifies visual question answering challenge on a restaurant menu image, e.g., computing the total price of a margherita pizza and a calzone. Similarly, Figure~\ref{fig:video_overview} shows visual question answering on a dance video, posing intricate visual understanding and reasoning challenges for MLLMs.

Despite numerous attempts~\cite{yin2023survey}, current MLLMs still face challenges. Two main approaches emerge in a zero-shot setting. One enhances MLLM pre-training for comprehensive image and video understanding, but models like GPT-4V~\cite{achiam2023gpt4}, Gemini~\cite{team2023gemini}, Claude~\cite{TheC3}, BLIP~\cite{li2022blip}, and Intervid~\cite{wang2024internvid} struggle with spatial reasoning, diagrams, text in images, and complex spatio-temporal dependencies in long-form videos~\cite{2023opencompass}\cite{liu2024world}. Alternatively, augmenting MLLMs with external tools/models like HuggingGPT~\cite{shen2024hugginggpt}, AssistGPT~\cite{gao2023assistgpt}, VideoAgent~\cite{wang2024videoagent}, and MM-React~\cite{yang2023mmreact} aims to improve visual comprehension. However, determining appropriate tools and building pipelines for complex VQA tasks on both images and long-form videos remains challenging.

In this work, we build upon these emerging strategies while drawing inspiration from human cognitive processes in complex visual reasoning tasks. Humans typically employ an iterative process involving analysis, observation, evaluation, reasoning, and verification to arrive at an answer. For example, when faced with a restaurant menu image, humans thoroughly understand the information, identify ordered items and their prices, and then calculate the total. Similarly, in long-form video VQA, humans analyze the full video and its transcript, identify relevant clips, select pertinent frames for additional insight, and integrate all information to answer questions. Finally, they verify steps and reasoning to validate the answer. This iterative approach, known as Critical Thinking~\cite{wikicritical}, is a fundamental cognitive skill for making informed decisions and solving complex problems.

Inspired by human cognitive processes and critical thinking, we present \mmct, a multi-modal critical thinking agent framework for comprehensive visual understanding and reasoning. Our framework comprises of three components, \textbf{dynamic planning and reasoning},\textbf{ tool augmentation}, and \textbf{a vision-based critic}. Just like humans break down complex problems into manageable tasks, the dynamic planner in \mmct decomposes user queries and devises problem-solving strategies. Iteratively, it assesses the current reasoning process and determines necessary actions to thoroughly analyze multi-modal information. To overcome MLLM limitations, \mmct leverages external tools to gather extra information, akin to how we seek additional insights to make informed decisions. Once enough data is gathered, the iterative process concludes with providing an answer. Critical thinking involves verifying the final answer and self-reflection. Hence, we propose a novel vision-based critic component that evaluates evidence and assumptions analyzing both textual and multi-modal data. The critic component introduces a generic approach to automatically determine evaluation criteria based on task description and human intent, ensuring precise assessment of answer accuracy and reasoning coherence. Finally, the critic evaluates against the derived criteria to determine the accuracy of the answer and provides feedback to enhance the reasoning process, aiding in defining new plans based on current information. Figure~\ref{fig:image_overview} and~\ref{fig:video_overview} illustrates the workings of \mmct. 

Our work distinguishes itself in several key areas. Firstly, while previous methods like MM-REACT~\cite{yang2023mmreact}, HuggingGPT~\cite{shen2024hugginggpt}, AssistGPT~\cite{gao2023assistgpt}, and ViperGPT~\cite{suris2023vipergpt} excel at task breakdown and reasoning, they lack comprehensive planning across modalities and dynamic reasoning. Secondly, while these approaches focus solely on reasoning, they neglect verification and self-reflection. To address this, we present a novel approach that introduces a vision-based critic and the criteria for evaluation in a generic manner, leveraging insights from textual QA verifiers~\cite{ji-etal-2023-towards} to enhance reasoning. Thirdly, our framework is generic, applicable to both images and long-form videos across domains and datasets. Importantly, \mmct is modular, enabling easy integration of improvements from newer multi-modal models and foundational tools complementing their advancements.


The \mmct framework integrates over 20 tools for various vision tasks spanning image, video, audio, and textual understanding. Augmented with the planner \& reasoning and a vision-based critic, \mmct excels in solving real-world complex visual reasoning tasks. Through rigorous evaluations across image and video understanding benchmarks, we demonstrate that \mmct (with and without critic) outperforms both foundational MLLMs and other tool-augmented pipelines. Notably, on image understanding datasets such as MMMU~\cite{yue2023mmmu}, MMVET~\cite{yu2023mmvet}, and MathVista~\cite{lu2024mathvista}, as well as MMBench~\cite{liu2024mmbench} and OKVQA~\cite{marino2019okvqa}, \mmct achieves exceptional performance, surpassing current state-of-the-art foundational models and approaches by \textbf{$10$\%}. For video QA, we evaluate on EgoSchema~\cite{mangalam2023egoschema}, a well-established dataset, and introduce a new dataset -- MMCT-QA -- comprising of 129 QA pairs across six distinct categories. \mmct achieves $71.2$\% accuracy on EgoSchema, outperforming state-of-the-art approaches by \textbf{$10$\%}, showcasing its effectiveness in tackling complex visual reasoning problems. 
To summarize, our key contributions are as follows: 
\begin{itemize}[noitemsep,topsep=0pt]
\item \textbf{\mmct:} A generic, agent-based multi-modal framework inspired by human cognitive and critical thinking process, for complex visual reasoning on images and long-form videos.

\item \textbf{Novel Vision-based Critic:} Within \mmct, we introduce a vision-based critic that autonomously identifies task-specific evaluation criteria and provides feedback. This enhances decision-making by integrating verification and self-reflection mechanisms.

\item \textbf{Comprehensive Evaluations and Analysis:} Through rigorous evaluations and ablation analyses across diverse image and video benchmarks, we showcase the robustness and effectiveness of \mmct, comparing it against end-to-end MLLMs and other pipelines.
\end{itemize}

\vspace{-12pt}
\section{Related Work}
\label{sec:rw}
\vspace{-8pt}
\textbf{Tool-Augmented Pipelines for Planning and Reasoning:} Tool-augmented LLM pipelines tackle MLLM limitations in multi-modal understanding, task comprehension, and reasoning. Examples like Chain-of-Thought Prompts~\cite{wei2023chainofthought}, Toolformer~\cite{schick2024toolformer}, and ReAct~\cite{yao2023react} showcase LLMs' role in problem-solving. MMReact~\cite{yang2023mmreact} extends ReAct for multi-modal systems, enabling problem breakdown and action planning. HuggingGPT~\cite{shen2024hugginggpt} breaks down user queries into sub-tasks, assigning vision models via a selection algorithm. The Chameleon~\cite{lu2024chameleon} pipeline adapts tools and domain expert models based on the LLM query planner. ViperGPT~\cite{suris2023vipergpt} and AssistGPT~\cite{gao2023assistgpt} provide visual interpretation through Python program execution. In contrast, \mmct employs a human-inspired critical thinking framework for task decomposition, strategy planning, and dynamic reasoning, enhancing decision-making by summarizing intermediate information, unlike static reasoning flows.

\textbf{Verification and Self-Reflection for MLLMs:}
Recently, Large Language Models (LLMs) have been used as verifiers across various tasks~\cite{weng2023large}. Typically, an LLM is queried for an answer and then re-queried with its response for critique or improvement, mostly focusing on NLP problems~\cite{shinn2023reflexion}.  For example, AssistGPT~\cite{gao2023assistgpt} includes a learner module that verifies the final answer with ground-truth samples, operating at a textual level and using them as in-context examples. In contrast, \mmct uses the entire reasoning chain and multi-modal data (e.g., images, videos) for verification and self-reflection, operating in a zero-shot manner. Other approaches like IdealGPT~\cite{you2023idealgpt} provides thorough reasoning for or against the proposed answer. IPVR~\cite{chen2023seethinkconfirm} decomposes the VQA task into phases, using an LLM to generate rationales in the ``confirm" module. Identifying the right criteria to evaluate against remains still a key challenge. Our work reformulates the critic definition for MLLMs, offering a novel approach to automatically define a vision-based critic, identify task-specific evaluation criteria, and provide structured self-reflection. This automated critic approach allows seamless integration of task-specific evaluation criteria into existing LLM pipelines.

\textbf{Long-Form Video Understanding:}
Long-form video understanding is challenging due to LLM limitations in handling long contexts and processing visual information efficiently~\cite{lin2023learning}. Foundational MLLMs like GPT-4V, Gemini, and Claude struggle with context length, with Azure GPT-4V API supporting only 10 frames per call~\cite{Microsoft2024}. A common approach is subsampling videos and passing each chunk to an MLLM for description, as shown by LLoVi~\cite{zhang2024simple}, which captions video clips and prompts an LLM with these captions. Alternatively, MLLM pipelines like VideoAgent~\cite{wang2024videoagent} and its extensions~\cite{fan2024videoagent} perform iterative VQA on videos by sampling the video linearly, generating descriptions using vision language models, and using tools like CLIP~\cite{radford2021learning} for iterative visual information retrieval. This method is expensive and inefficient and varies by dataset. Our framework differs by using a generic iterative approach. It first indexes the entire video using tools like Azure Video Retriever~\cite{Microsoft2024ComputerVision} or CLIP-based models~\cite{yuan2021florence}. Then, it uses transcripts (where available) to determine relevant time frames or rewrites the user query with a planner and reasoner to retrieve frames of interest, which are analyzed by MLLMs like GPT-4V. \mmct combines planning and reasoning with external tools, including MLLMs and vision models, for comprehensive video analysis. Our critic verifies the answer by analyzing the entire reasoning chain and selected video clips through visual analysis.


\begin{figure}
    \centering
    \includegraphics[width=0.85\textwidth]{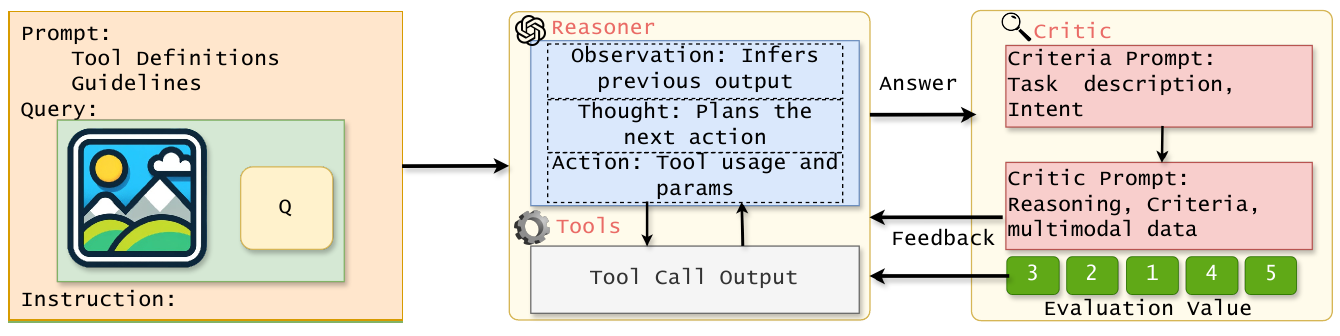}
    \vspace{-10pt}
    \caption{\mmct Overview with Planner and Reasoner, Tools, and Critic Components.}
    \label{fig:mmct_overview}
    \vspace{-15pt}
\end{figure}
\vspace{-10pt}
\section{\mmct~Overview}
\label{sec:overview}
\vspace{-8pt}
\mmct is one of the first generic solution for complex visual understanding and reasoning tasks, applicable to both images and long-form videos. It processes user queries with images or videos to generate precise answers grounded in multi-modal information. Inspired by human critical thinking, \mmct adopts an iterative approach for detailed analysis, reasoning, information gathering, and answer verification. At the core of the \mmct framework are three key components: \textit{the dynamic planner and reasoner, the tool augmentation, and the vision-based critic}, described next.
\vspace{-10pt}
\subsection{Dynamic Planner and Reasoner }
\label{sec:planner}
\vspace{-2pt}
The planner and reasoner serves as the central orchestrator of \mmct. It breaks down user queries into sub-tasks, creates problem-solving strategies, and adapts based on new information. Leveraging the high-level planning abilities of LLMs and the ReAct~\cite{yao2023react} framework for reasoning, \mmct~efficiently solves complex visual tasks.
Input to the reasoner includes a \texttt{[problem description]} providing a high-level task overview, \texttt{[instructions]} detailing the critical thinking approach to solving it, \texttt{[tool descriptions]} listing available tools and their functionalities, \texttt{[user query]} defining the question of interest, and \texttt{[multi-modal data]} such as images, videos. (See Appendix~\ref{app:planner_prompt} for sample prompts).

Initially, the reasoner uses a vision interpreter tool such as a MLLM or vision model, to gather comprehensive information about the multi-modal data, aiding in planning and reasoning. Using this information, along with the problem description and user query, the reasoner formulates a plan and associated reasoning. Guided by the instructions, it dynamically generates a plan, identifies the next step, acquires additional information, and iteratively updates the plan and reasoning. Each step involves a \textit{thought} (assessing relevant evidence, observations, and reasoning for potential next steps), an \textit{action} (acquiring more information), and an \textit{observation} (analyzing the information gained). Unlike static approaches, the dynamic planner and reasoner continuously evaluate the current reasoning process and adjust actions accordingly. This adaptability enables the system to excel in multi-modal understanding and reasoning, ensuring effectiveness in handling complex tasks.
\vspace{-8pt}
\subsection{Tool Augmentation}
\label{sec:tools}
\vspace{-5pt}
This component enables seamless integration of various general-purpose or domain-specific tools, empowering it to gain additional insights from multi-modal data. Equipped with descriptions and metadata of these tools, \mmct dynamically invokes them during its critical thinking process. We leverage the following tools to attain a comprehensive understanding of multi-modal data:

\textbf{1. Image Understanding \& Descriptors:} These tools specialize in interpreting visual content within images.
\textit{(a) VIT (Vision Interpreter):} VIT aids in image classification and understanding, extracting high-level visual features for tasks like object recognition, scene understanding, etc. 
\textit{(b) OCR (Optical Character Recognition):} OCR extracts text from images. 
\textit{(c) Object Detection:} Object detection identifies and localizes objects within images. 
\textit{(d) Recognition (Face/Object Recognition): }Recognition identifies specific objects or faces within images. 
Appendix.~\ref{app:tools} provides details of the exact tools supported by \mmct. 

\textbf{2. Audio Analysis \& Descriptors:} ASR (Automatic Speech Recognition) is utilized to transcribe spoken language into text, essential for tasks like audio data transcription and multi-modal analysis. 

\textbf{3. Textual Analysis \& Retrievers:}
This tool retrieves semantically matched phrases from transcripts based on a search query, aiding tasks like retrieval and context understanding. It employs embedding models to encode phrases and search queries, returning top matches using cosine similarity.

\textbf{4. Video Analysis \& Retrievers:}
This tool analyzes video frames to create a queriable index, using video embeddings like CLIP~\cite{radford2021learning} to identify specific moments. It aids in tasks such as video summarization and  analysis, enhancing the understanding of visual information within videos. 

\textbf{5. Video Understanding \& Descriptors:} This tool utilizes foundational models (MLLMs) to analyze multiple video frames simultaneously, enabling comprehensive multi-modal analysis. 

Note that, the current set of tools added are quite generic and works for various tasks and domains, furthermore additional tools can be added seamlessly as required (See Appendix~\ref{app:tools} for more details).
\begin{figure}
    \centering
    \includegraphics[width=\textwidth]{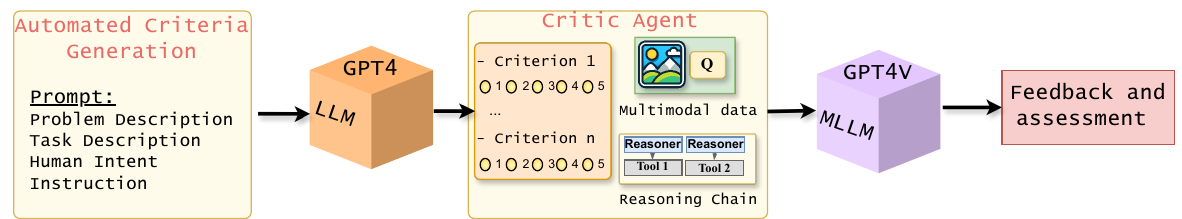}
    \vspace{-5pt}
    \caption{Vision-based Critic overview.}
    \label{fig:critic}
    \vspace{-15pt}
\end{figure}
\vspace{-12pt}
\subsection{Vision-based Critic}
\label{sec:critic}
A crucial component of critical thinking is verifying the final answer and engaging in self-reflection. We introduce a novel vision-based critic using an MLLM like GPT-4V, which scrutinizes the reasoning chain, including evidence, assumptions, and accompanying image or video data. Unlike textual critics that focus solely on reasoning, this vision-based approach analyzes all information, ensuring robust verification and self-reflection.
Previous works with LLMs for verification lacked explicit evaluation criteria, limiting their effectiveness \cite{zhang2023llmeval}. Users had to establish these criteria themselves, which was burdensome due to task diversity. Our approach automates the definition of evaluation criteria upfront using LLMs (see Figure~\ref{fig:critic}). These criteria are integrated into the vision-based critic, enabling it to assess the final answer and offer constructive feedback.

To automatically identify task-specific evaluation criteria, we use an LLM like GPT-4, processing inputs such as \texttt{[Problem description]}, \texttt{[Instruction]}, \texttt{[Task Description]}, and \texttt{[Human Intent]}. The problem description provides an overview, instructions detail how to define criteria, the task description offers specific information (e.g., VQA), and human intent specifies qualitative metrics (e.g., concise answers, clear reasoning). Using this input, the LLM formulates criteria, descriptions, and acceptable values. Example criteria derived, \textbf{Criteria}: Clarity of Reasoning, \textbf{Description}: Logic behind the model's answer, demonstrating its understanding, \textbf{Acceptable Values}: "1": "Not clear", "2": "Somewhat clear", "3": "Clear", "4": "Very clear", "5": "Extremely clear" (See Appendix~\ref{app:critic} for more details and prompts).  

The vision-based critic uses task-specific criteria to systematically evaluate the reasoning chain, evidence, and multi-modal data. Inputs include \texttt{[Problem description]}, \texttt{[Instructions]}, \texttt{[Evaluation results]}, and \texttt{[Feedback]}. The problem description outlines the critic's task, instructions specify evaluation methods, evaluation results prompt specific output formats, and feedback guides accurate reasoning or self-reflection. This helps the planner and reasoner determine next steps. The critic is invoked only after the final answer, as experimenting with the critic at individual steps showed no performance improvement due to limited knowledge at each step. 

We use GPT-4V from Azure OpenAI~\cite{Microsoft2024Azure} as our critic model for image and video understanding. While the critic can process an entire image, it faces constraints with videos, only handling 10 frames at a time due to API constraints~\cite{Microsoft2024}. To work around this, for video comprehension, we pick the top-3 relevant video clips. Next, we extract frames from these clips, creating image sets (max 10) resembling a photo grid of size nxn (where n is the number of frames concatenated in a image). These image sets are then fed to the vision-based critic for comprehensive analysis and evaluation (see Appendix~\ref{app:photogrid} for examples).

\vspace{-10pt}
\section{Qualitative Examples}
\label{sec:examples}
\vspace{-5pt}
\begin{figure*}[!t]
\begin{minipage}[t]{0.5\linewidth}
        \includegraphics[trim=1.5cm 0cm 1.4cm 0cm, clip,width=1.0\columnwidth]{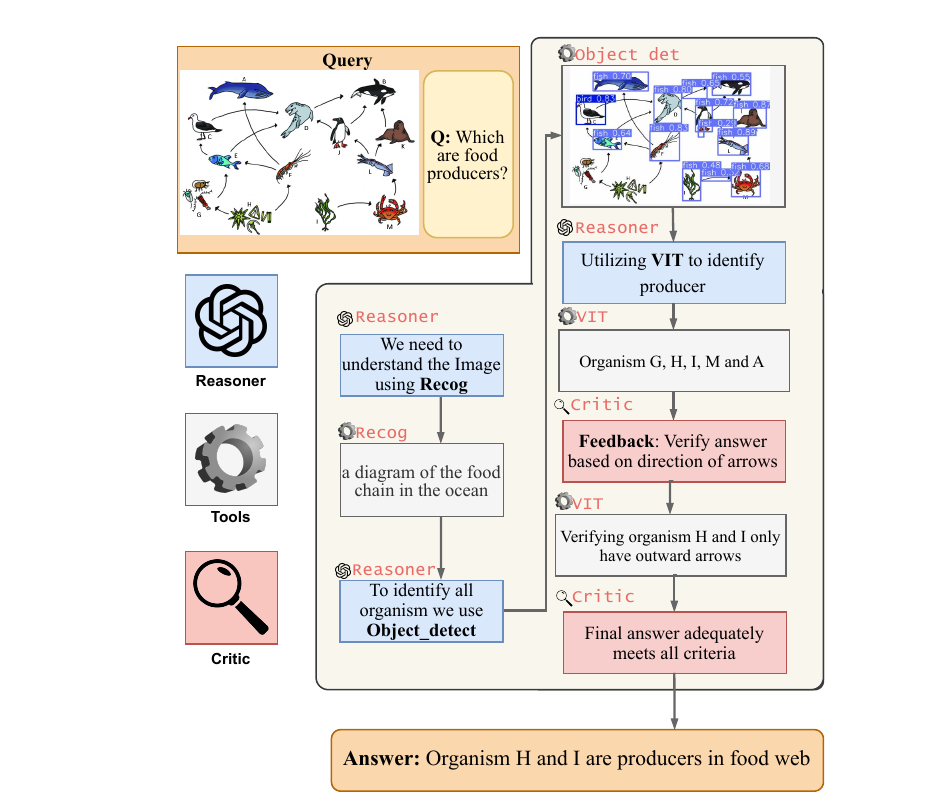}
        \caption{Image understanding and reasoning.}
        \label{fig:image_example}
        \vspace{-5pt}
\end{minipage}
\hspace{3pt}
\begin{minipage}[t]{0.5\linewidth}
        \includegraphics[width=1.0\columnwidth ]{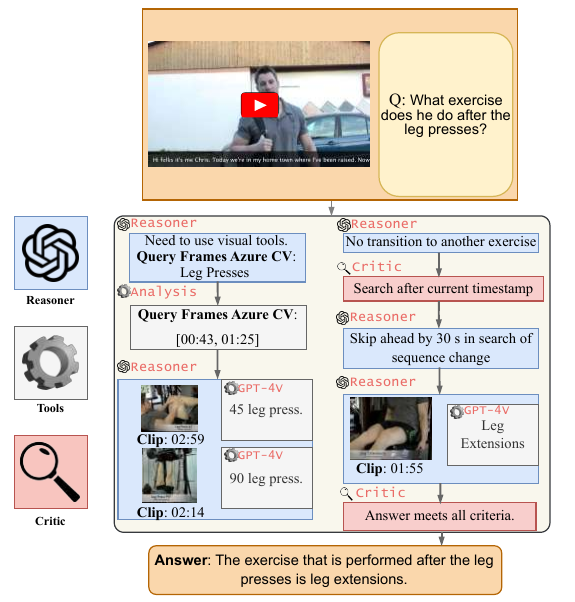}
        \caption{Video understanding and reasoning.}
        \label{fig:video_example_main}
        \vspace{-5pt}
\end{minipage}
        \vspace{-15pt}
\end{figure*}
\textbf{Image understanding and reasoning.}
Figure~\ref{fig:image_example} illustrates \mmct's execution with an image, tackling the user query ``which are the producers in this food web?" Foundational MLLMs like GPT-4V and Gemini yield incorrect answers, emphasizing the complexity of the task. \mmct begins with an initial analysis using a Vision Transformer (VIT) tool, \textit{recognizing} the image as an ocean food chain diagram. Subsequent steps involve \textit{object detection} to identify organisms and \textit{MLLM analysis} to pinpoint independent organisms within the food web. The planner integrates insights from the MLLM and previous observations to derive a preliminary answer. The critic then evaluates the answer, highlighting \textit{deficiencies in comprehensiveness and clarity}, leading to a refined reasoning chain focusing on directional aspects of arrows between organisms. The revised answer undergoes validation by the critic, confirming its accuracy and coherent reasoning. This example showcases \mmct's proficiency in complex visual reasoning, facilitated by iterative critical thinking involving the planner \& reasoner, tool augmentation, and vision-based critic (see Appendix\ref{app:image_quali} for more examples).

\textbf{Video understanding and reasoning.}
Figure~\ref{fig:video_example_main} illustrates \mmct's analysis of a fitness video, focusing on identifying the exercise after leg presses. Despite context length limitations, \mmct overcomes this challenge with its generic approach. The process begins with initial assessment and \textit{audio transcription}, while the video undergoes \textit{visual indexing}. Transcript analysis reveals no direct references, prompting \textit{visual tools' use} to identify relevant clips. Employing Azure Computer Vision, \mmct identifies leg press timestamps but finds no subsequent exercise. GPT-4V analysis continues to show only leg presses. Following \textit{critic feedback}, the search extends to frames 30 seconds post timestamps, revealing \textit{leg extensions}. Integrating these insights, \mmct confirms leg extensions as the subsequent exercise. The \textit{critic evaluates} the final answer, confirming its accuracy and reasoning clarity. This robust analysis, emphasizing critical thinking, overcomes information gaps using advanced visual tools (see Appendix\ref{app:video_quali} for more examples).

\vspace{-12pt}
\section{Datasets and Metrics}
\label{sec:datasets}
\vspace{-10pt}
We conduct rigorous evaluations across image and video understanding benchmarks in a zero-shot setting. Our evaluation assesses \mmct's ability in multi-modal understanding and reasoning, integrating visual and textual data, applying domain-specific knowledge, and utilizing external information. The evaluation metric for all datasets is the accuracy of answers to all questions.

\textbf{Image understanding benchmarks.} 
We evaluate \mmct across five challenging and diverse image datasets, \texttt{MMVET}~\cite{yu2023mmvet}, \texttt{MMMU}~\cite{yue2023mmmu}, \texttt{MMBench}~\cite{liu2024mmbench}, \texttt{OKVQA}~\cite{marino2019okvqa},   \texttt{MathVista}~\cite{lu2024mathvista}, each with its unique focus and challenges, to comprehensively assess its capabilities in multi-modal understanding and reasoning. More details of the datasets are in Appendix~\ref{app:mmct_data}.

\textbf{Video understanding benchmarks.} We evaluate \mmct on a widely recognized long-form video dataset EgoSchema and introduce our own dataset for complex reasoning and video analysis.

\underline{\texttt{Egoschema}}~\cite{mangalam2023egoschema} consists of 5000 multiple-choice questions sourced from 5000 egocentric videos covering a wide array of natural human activities. Each video spans 3 minutes, and the dataset comprises a test set, with a subset of 500 questions having publicly available labels.


\underline{\texttt{MMCT-QA}} aims to create a benchmark for video understanding that meets three criteria: (i) representation of long-form videos, (ii) realism in scenarios requiring different capabilities, and (iii) inclusion of both audio and video modalities. We structured a taxonomy of queries into six categories: temporal understanding, spatial understanding, event \& action recognition, dialogue \& transcript-based, abstract and conceptual, and specific detail based, each targeting different video understanding aspects. Our dataset includes 15 diverse videos sourced from the Youtube 8M~\cite{abuelhaija2016youtube8m} dataset (which we modify and distribute under its Apache License 2.0) with 129 question-answer pairs, created by human annotators. 
Since the answers are open-ended, an LLM-based evaluator verifies system-generated answers against ground truth, categorizing them as no match, partial match, or complete match (see Appendix~\ref{app:mmct_data} for sample data).

\textbf{Implementation Details.} \mmct uses the same configurations for both image and video analysis across datasets. The planner and reasoning agent use GPT-4 (gpt-4-32k (0613)~\cite{openai2023gpt4blog}) as the LLM for all experiments. The VIT tool and the critic used is a MLLM, i.e., GPT-4v (gpt-4 (vision-preview)). Note that all our evaluations are in a zero-shot setup, unlike AssistGPT~\cite{gao2023assistgpt}. The source code for \mmct and the MMCT-QA dataset is available for the community\footnote{The source code will be released soon. }. To run the pipeline we use a Virtual Machine composed of 1 x A100 80 GB, 64 cpu cores at 3.2GHz and 512 GB RAM. GPU is necessary to support tools at that are inferred locally.
\vspace{-10pt}
\section{Results: Image Understanding and Reasoning}
\label{sec:image_results}
\vspace{-5pt}
We meticulously evaluate \mmct against established benchmarks, including foundational MLLMs like GPT-4V~\cite{achiam2023gpt4}, Claude~\cite{TheC3}, Gemini~\cite{team2023gemini}, and other tool-based MLLMs like AssistGPT~\cite{gao2023assistgpt} and ViperGPT~\cite{suris2023vipergpt}, to assess its effectiveness. 
\begin{table}
\centering
\resizebox{\textwidth}{!}{%
\begin{tabular}{llllllllll}
\hline
Dataset &
  \begin{tabular}[c]{@{}l@{}}Claude 3\\ Opus*\end{tabular} &
  \begin{tabular}[c]{@{}l@{}}Claude 3\\ Sonnet*\end{tabular} &
  \begin{tabular}[c]{@{}l@{}}Claude 3\\ Haiku*\end{tabular} &
  GPT-4V* &
  \begin{tabular}[c]{@{}l@{}}Gemini\\ 1.0 Ultra*\end{tabular} &
  \begin{tabular}[c]{@{}l@{}}Gemini\\ 1.5 Pro*\end{tabular} &
  \begin{tabular}[c]{@{}l@{}}Gemini\\ 1.0 Pro*\end{tabular} &
  \begin{tabular}[c]{@{}l@{}}MMCT\\ w/o Critic\end{tabular} &
  \begin{tabular}[c]{@{}l@{}}MMCT \\ w Critic\end{tabular} \\ \hline
MMMU      & 59.40 & 53.10 & 50.20 & 56.80 &59.40 & 58.50 & 47.90 & \underline{59.54} & \textbf{63.57} \\
MathVista & 50.50 & 47.90 & 46.40 & 49.90 & 53.00 & 52.10 & 45.20 & \underline{53.30} & \textbf{56.50} \\
MMVET     & 51.70 & 51.30 & -     & 60.20 & -     & 64.20 & -     & \underline{70.51} & \textbf{74.24} \\
MMBench   & 63.30 & 67.80 & 60.70 & 77.00 & -     & 73.60 & -     & \underline{80.21} & \textbf{84.20}\\ \hline
\end{tabular}%
}
\caption{\mmct outperforms SOTA foundational models across all datasets (Bold: best, Underline: second best). * Sourced directly from original reports. }
\label{tab:img_perf}
\vspace{-18pt}
\end{table}
\vspace{-5pt}
\subsection{Performance analysis}
Table~\ref{tab:img_perf} presents the performance analysis of \mmct compared to state-of-the-art (SOTA) MLLMs across all datasets. \mmct, equipped with a vision-based critic, consistently outperforms SOTA MLLMs such as Claude 3, GPT-4V, and Gemini models by at least \textbf{10\% across all datasets}. For instance, on the MMVET dataset, \mmct achieves 74.2\% accuracy, showcasing performance improvement by +22.3\%, +14.1\%, and +10.4\% points over Claude 3, GPT-4V, and Gemini models, respectively. This trend persists across all datasets, with \mmct on average surpassing GPT-4V by 10\%, Claude 3 by 15\%, and Gemini models by 10\%. This performance enhancement highlights the synergy among the three proposed components within \mmct, enabling comprehensive analysis of image data. The performance boost can be attributed to several factors within our pipeline: 1) Utilization of superior tools for individual capabilities compared to the inherent capabilities of MLLMs, 2) Implementation of an iterative reasoning chain that decomposes tasks into manageable subtasks, and 3) Integration of a vision-based critic for thorough evaluation of derived answers, reasoning chains, and multimodal data. It's noteworthy that even without the critic, \mmct outperforms all SOTA MLLMs (second best- underlined). 

Furthermore, Figure~\ref{tab:img_tools_perf} shows \mmct's superior performance to SOTA tool-based approaches like AssistGPT~\cite{gao2023assistgpt} and ViperGPT~\cite{suris2023vipergpt} on OKVQA dataset. We select OKVQA as this was the only dataset other tool-based models were evaluated. \mmct outperforms AssistGPT by 12\% and ViperGPT by 5\%, respectively. 

\texttt{Summary:} \textit{\mmct outperforms all SOTA MLLMs and tool-based MLLMs across several challenging image benchmarks by atleast 10\%.}


\subsection{Vision-based critic performance}
\label{sec:critic_res_img}
\begin{figure*}[!t]
\begin{minipage}[t]{0.3\linewidth}
        \includegraphics[width=1\linewidth, height=1in]{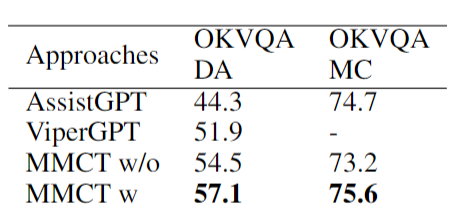}
        \caption{\mmct performance on OKVQA.}
        \label{tab:img_tools_perf}
        \vspace{-5pt}
\end{minipage}
~
\begin{minipage}[t]{0.21\linewidth}
        \includegraphics[trim=0.5cm 0.5cm 1.5cm 0.8cm, 
 clip, width=\linewidth]{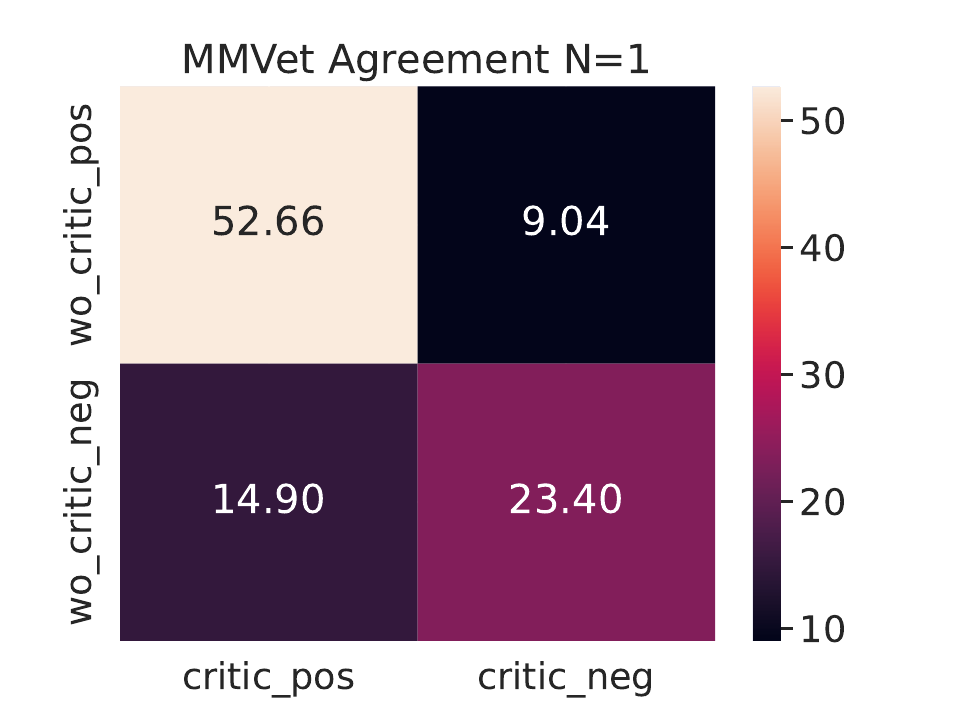}
        \caption{Confusion matrix (N=1).}
        \label{fig:cf1}
        \vspace{-5pt}
\end{minipage}
\begin{minipage}[t]{0.21\linewidth}
        \includegraphics[ trim=0.5cm 0.5cm 1.5cm 0.8cm, clip, width=\linewidth]{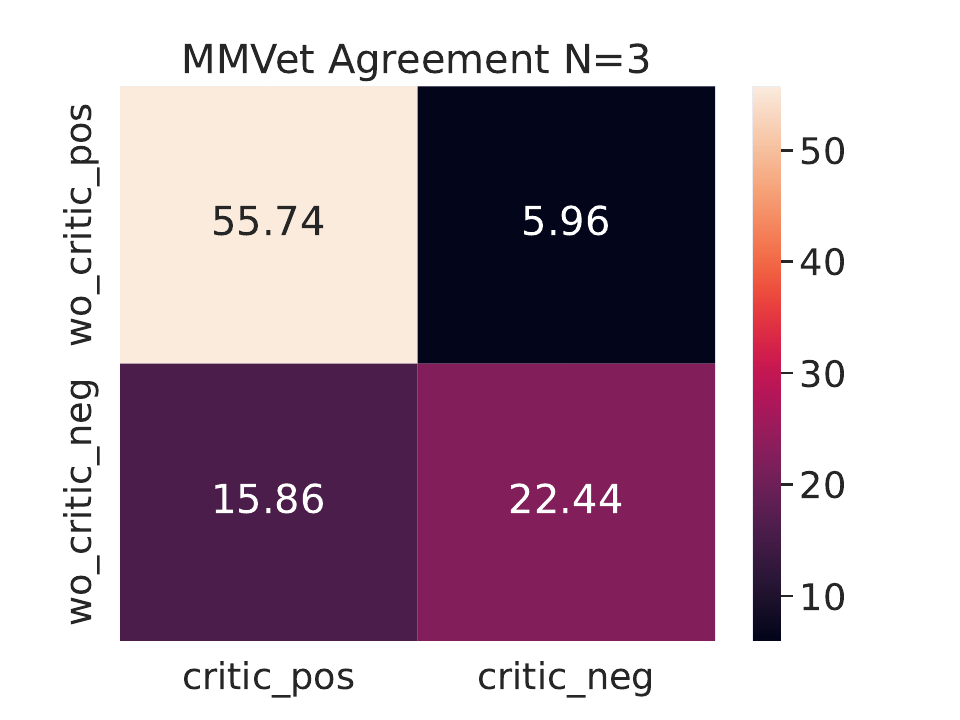}
        \caption{Confusion matrix (N=3).}
        \label{fig:cf2}
        \vspace{-5pt}
\end{minipage}
\begin{minipage}[t]{0.21\linewidth}
        \includegraphics[ trim=0.5cm 0.5cm 1.5cm 0.8cm, clip, width=\linewidth]{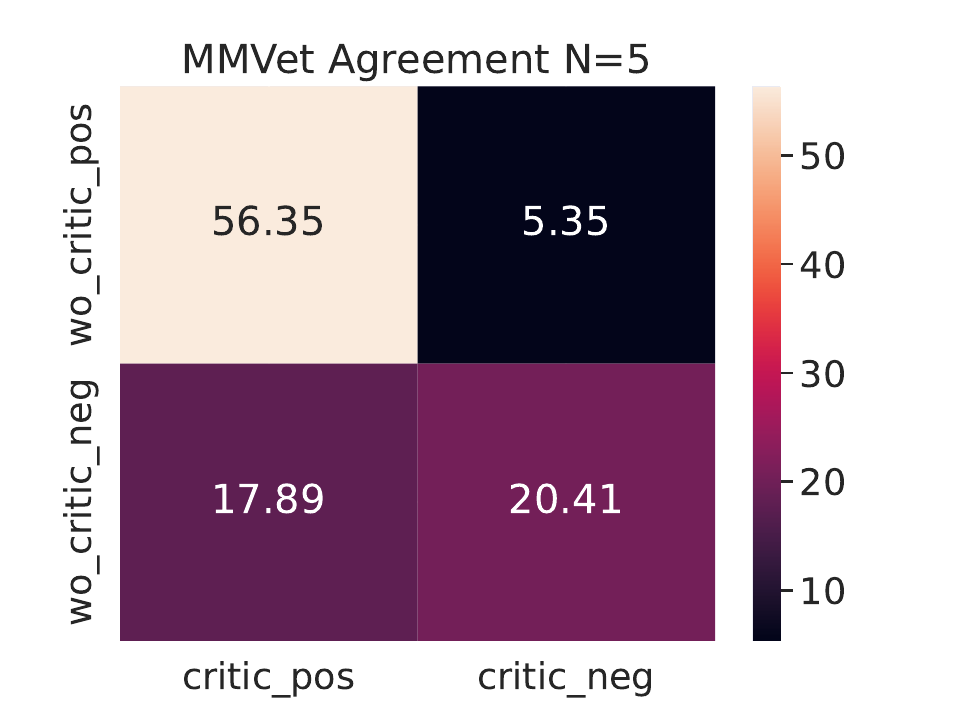}
        \caption{Confusion matrix (N=5).}
        \label{fig:cf3}
        \vspace{-5pt}
\end{minipage}
\vspace{-5pt}
        \vspace{-12pt}
\end{figure*}

We now assess the effectiveness of the vision-based critic within \mmct. From Table~\ref{tab:img_perf} we can see that by introducing critic, \mmct's performance \textbf{improves on average by 5\%}. Figures~\ref{fig:cf1},~\ref{fig:cf2}, and~\ref{fig:cf3} show the confusion matrix of \mmct with and without critic for MMVET dataset. Each cell denotes the \% of samples that fall under the criteria. The top-left cell signifies cases where both, with and without the critic, agree. The top-right cell indicates samples where \mmct without the critic produced the correct answer, but the introduction of the critic led to incorrect answers. Conversely, the bottom-left cell indicates instances where the introduction of the critic assisted \mmct in deriving correct answers that were not feasible earlier. Finally, the bottom-right cell depicts samples where neither approach was able to derive the correct answer.

\mmct, with and without the critic, agrees on 52-56\% of samples. Here, the critic verifies and solidifies answers but doesn't improve performance. In 20-23\% of samples, neither approach could derive the correct answer, mainly due to limitations in the tools and MLLMs' comprehension abilities. The introduction of the critic generally enhances performance in 14-18\% of samples. However, there are instances where it does not perform optimally: (1) When the base pipeline suggests an incorrect answer and the critic accepts it, and (2) When the pipeline would have arrived at the correct answer, but the critic leads it to select the wrong answer.

\texttt{Base framework derives wrong answer and Critic accepts it:} The primary reason for these issues is that \mmct uses GPT-4V for both the VIT and critic, resulting in shared weaknesses. To mitigate this, employing different MLLMs for these roles could be effective. 

\texttt{Critic leads the base framework to the wrong answer:} There are few instances where the critic results in the wrong answer. These occur when the base framework had the correct answer, but the critic's attempt to extract additional information leads to hallucinations. Currently, the critic prioritizes specificity over simplicity. To mitigate such errors, establishing more detailed guidelines and enhancing the critic evaluation criteria could be effective.

We conduct experiments with varying numbers of critic calls (N=1, 3, and 5) to understand the critic's impact. Figures~\ref{fig:cf1},~\ref{fig:cf2}, and~\ref{fig:cf3} show that increasing critic calls from 1 to 3 reduces adversarial effects (disagreement drops from 9\% to 5\%). However, further increasing the number of calls beyond 3 does not significantly boost performance. We observe a notable improvement of 14.75\% with one critic call, while additional calls result in only minimal increase in accuracy. Due to space constraints, we do not provide an in-depth critic analysis for other datasets, but they follow similar trends. Appendix~\ref{app:critic_quali} offers additional details and qualitative examples for these scenarios.

Summary: \textit{\mmct with critic boosts the performance by 5\% and also assists in validating and grounding the generated answer. }









 


\vspace{-10pt}
\section{Results: Video Understanding and Reasoning}
\label{sec:video_results}

\begin{wraptable}{R}{4cm}
\vspace{-15pt}
\resizebox{0.3\textwidth}{!}{%
\begin{tabular}{ll}
\hline
Method          & Acc. \\ \hline
GPT-4V~\cite{achiam2023gpt4}           & 63.5     \\
Gemini 1.0 Pro~\cite{team2023gemini}   & 61.5     \\ \hline
LLoVi~\cite{zhang2024simple}            & 57.6     \\
MC-ViT-L~\cite{balažević2024memory}         & 62.6     \\
Video-LLaVa~\cite{lin2023videollava}      & 36.8     \\
ViperGPT~\cite{suris2023vipergpt}        & 15.8     \\
VideoAgent~\cite{wang2024videoagent}       & 60.2     \\
VideoAgent-M~\cite{fan2024videoagent}    & 62.8     \\ \hline
MMCT w/o Critic & 68.8     \\
MMCT w Critic   & \textbf{71.2}     \\ \hline
\end{tabular}%
}
\caption{\mmct on EgoSchema.}
\label{tab:vid-perf}
\vspace{-14pt}
\end{wraptable}

\vspace{-5pt}

\subsection{Performance Analysis}
Table~\ref{tab:vid-perf} demonstrates \mmct's superiority over all state-of-the-art (SOTA) methods. On the EgoSchema~\cite{mangalam2023egoschema} 500-question subset (3-minute, no-audio videos), \mmct achieves an accuracy of 71.2\% with a critic and 68.8\% without one, surpassing models like LLoVi, Video-LLaVa, MC-ViT-L, VideoAgent, VideoAgent-M, ViperGPT, GPT-4V, and Gemini 1.0 Pro. Notably, \mmct improves performance \textbf{by 10\% }over Gemini and GPT-4V, showcasing its effectiveness in complex visual reasoning tasks.

On our MMCT-QA dataset, \mmct outperforms standard baselines using GPT-4V by 20\% on average, as shown in Table~\ref{tab:mmct-perf}. Due to limitations in code availability, context length, and computational challenges, we compared \mmct against two baselines: 

\begin{wraptable}{R}{4cm}
\vspace{-15pt}
\resizebox{0.25\textwidth}{!}{%
\begin{tabular}{ll}
\hline
Method           & Accuracy                    \\ \hline
Baseline1        & { 41.1}   \\
Baseline2        & { 51.2}   \\
MMCT w/o critic  & { 67.1} \\
MMCT with critic & { \textbf{71.3}}
\end{tabular}%
}
\caption{\mmct on MMCT-QA.}
\label{tab:mmct-perf}
\vspace{-15pt}
\end{wraptable}

(i)\textit{ Baseline-1 (B1):} Videos are divided into five random 10-second clips. GPT-4V generates descriptions for each clip, which are aggregated with audio transcripts. GPT-4 then answers queries based on this textual description. (ii) \textit{Baseline-2 (B2):} Builds on B1 by embedding each clip's description. The closest chunk based on description embedding is retrieved and passed to GPT-4V to answer the query. B1 converts video and audio data to text for answering queries, similar to MMVid~\cite{lin2023mmvid}. B2 retrieves the best chunk for analysis, similar to AssistGPT~\cite{gao2023assistgpt}. \mmct outperforms both B1 and B2 on our dataset. 

\texttt{Summary:} \textit{\mmct, with and without a critic, achieves SOTA accuracy on both EgoSchema and MMCT-QA datasets, outperforming proprietary, public, and tool-based MLLMs. These results underscore \mmct's effectiveness and efficiency in tackling complex visual reasoning problems.}

\vspace{-5pt}
\subsection{Vision-based critic performance}

\begin{wrapfigure}{r}{4cm}
\vspace{-10pt}
    \centering
\includegraphics[width=0.25\columnwidth, height=1.2in]{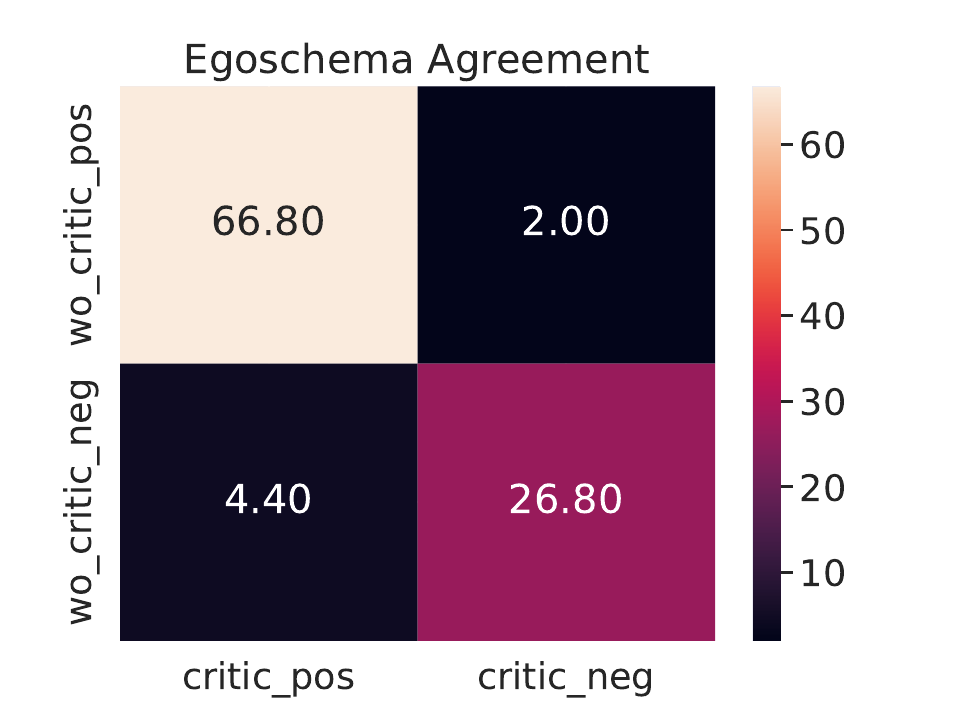}
        \caption{Confusion matrix. }
        \label{fig:cf_video}
        \vspace{-12pt}
\end{wrapfigure}
Similar to the critic analysis on image datasets in Section~\ref{sec:critic_res_img}, introducing the vision-based critic improves \mmct's performance by 3-4\% across both video benchmarks. This improvement is achieved by passing selected frames in a photo grid (image set) to the critic for verification and self-reflection, as detailed in Section~\ref{sec:critic}. Appendix~\ref{app:photogrid} provides further details and sample images of the multi-modal data used in this photo grid manner for the critic.

Figure~\ref{fig:cf_video} shows the confusion matrix of \mmct with and without the critic for the EgoSchema dataset. The introduction of the critic not only helps validate current answers (66\%) but also improves performance in cases where the default pipeline fails (5\%). 
  
\texttt{Summary:} \textit{\mmct with the critic significantly enhances long-form video comprehension, boasting an improvement of +4\% and ensuring grounded answers within the multi-modal data. }

\section{Conclusions}
\label{sec:conc}
In this work, we introduced \mmct, a novel multi-modal critical thinking agent framework designed to enhance visual reasoning capabilities in MLLMs. Inspired by human cognitive processes and critical thinking, \mmct addresses the limitations of current MLLMs in multi-modal processing and reasoning over complex visual tasks by integrating dynamic planning, tool augmentation, and a novel vision-based critic. The critic evaluates evidence and assumptions, determines answer accuracy, and provides feedback to enhance reasoning. Our performance analysis demonstrates that \mmct consistently outperforms state-of-the-art (SOTA) models like Claude 3, GPT-4V, and Gemini by at least 10\% across various image and video datasets, with the critic improving overall accuracy by 5\%. \mmct's modularity allows seamless integration of advancements in multi-modal models and tools, ensuring continuous improvements in visual reasoning. Furthermore, the framework's generic approach makes it applicable across various domains and datasets.
\textbf{Limitations:} Despite the use of the critic, \mmct can still hallucinate and generate incorrect answers; additional measures are necessary to verify the reasoning chain. While \mmct has shown promising results across various datasets, applying it to real-world scenarios requires further testing. Additionally, the dependency on external tools can introduce vulnerabilities if these tools fail or are unavailable, and the computational overhead of \mmct may limit real-time applicability.

\newpage

\bibliography{neurips_2024}
\bibliographystyle{abbrvnat}
\newpage

\section*{Appendix}

\section{Dynamic Planner and Reasoner Agent: Additional details}
\label{app:planner_prompt}
\definecolor{codegreen}{rgb}{0,0.6,0}
\definecolor{codegray}{rgb}{0.5,0.5,0.5}
\definecolor{codepurple}{rgb}{0.58,0,0.82}
\definecolor{backcolour}{rgb}{0.95,0.95,0.92}

\lstdefinestyle{mystyle}{
    backgroundcolor=\color{backcolour},   
    commentstyle=\color{codegreen},
    keywordstyle=\color{magenta},
    numberstyle=\tiny\color{codegray},
    stringstyle=\color{codepurple},
    basicstyle=\ttfamily\footnotesize,
    breakatwhitespace=false,         
    breaklines=true,                 
    captionpos=b,                    
    keepspaces=true,                 
    numbers=none,                    
    numbersep=5pt,                  
    showspaces=false,                
    showstringspaces=false,
    showtabs=false,                  
    tabsize=2,
    frame=single,                    
    rulecolor=\color{black},         
    framesep=5pt,                    
    framerule=0.5pt 
}

\definecolor{delim}{RGB}{20,105,176}
\definecolor{numb}{RGB}{106, 109, 32}
\definecolor{string}{rgb}{0.64,0.08,0.08}

\lstdefinelanguage{json}{
    numbers=left,
    numberstyle=\tiny\color{codegray},
    frame=single,
    rulecolor=\color{black},
    showspaces=false,
    showtabs=false,
    breaklines=true,
    postbreak=\raisebox{0ex}[0ex][0ex]{\ensuremath{\color{gray}\hookrightarrow\space}},
    breakatwhitespace=true,
    basicstyle=\ttfamily\small,
    upquote=true,
    morestring=[b]",
    stringstyle=\color{string},
    literate=
     *{0}{{{\color{numb}0}}}{1}
      {1}{{{\color{numb}1}}}{1}
      {2}{{{\color{numb}2}}}{1}
      {3}{{{\color{numb}3}}}{1}
      {4}{{{\color{numb}4}}}{1}
      {5}{{{\color{numb}5}}}{1}
      {6}{{{\color{numb}6}}}{1}
      {7}{{{\color{numb}7}}}{1}
      {8}{{{\color{numb}8}}}{1}
      {9}{{{\color{numb}9}}}{1}
      {\{}{{{\color{delim}{\{}}}}{1}
      {\}}{{{\color{delim}{\}}}}}{1}
      {[}{{{\color{delim}{[}}}}{1}
      {]}{{{\color{delim}{]}}}}{1},
}
\lstset{style=mystyle}

While both of our pipelines have the same functionalities, the prompt varies in terms of style and specific details. In this section, both prompts and structure are presented in a unified format.
\subsection{Image Pipeline}
\subsection*{Prompt Structure}
The prompt is structured using LLama\_Index, This is the primary library used in developed of the pipeline. We utilize a modified version of \texttt{ReactAgent} from LLama\_Index to enable easy integration and high control. The formatted prompt can be split into 3 sections, i.e., 1) Tool Descriptions, 2) Input-Output Definition, 3) Guidelines.

\subsubsection*{Tool Description}
\begin{lstlisting}[language=json, linerange={5-110}]
## Tools
You have access to a wide variety of tools. You are responsible for using
the tools in any sequence you deem appropriate to complete the task at hand.
This may require breaking the task into subtasks and using different tools
to complete each subtask.

You have access to the following tools:

> Tool Name: Vision Expert: vit
  Tool Description: You can query information about the given image/images using simple natural language,
                This returns responses in simple language.
                input: 
                    {"query": "What is the number of objects in the image"}
                    or 
                    {"query": "What is the number of objects in the image", "selected_image": "1"}

                    The input can contain two values "query" and "selected_image". "selected_image" is optional but "query" is necessary for all queries.
                    "query" is to define the question that the Vision expert would answer about the image.
                    "selected_image" is used only when there are multiple images given in the problem setting. There are three valid options for "selected_image" i.e., "1", "2", "all". By default all is used, and for scenarios where there is only one image "selected_image" do not change the selection of image.

                response:
                    The output is simple text answering the query given.

  Tool Args:  query: str
              selected_image: Optional[str] = "all" \n \t possible values: ["1","2",...(any number)...,"all"]

> Tool Name: Object Detection Tool: object_detect\n
  Tool Description: You can use this tool to analyze the given image, The tool should be used when
                individual objects are to be detected in the image. The algorithm returns
                positions of individual elements that it can detect.

                This returns response in a dictionary with the name of the object and the 
                position of the object in pixel coordinates in XYHW format.
                XYHW format represents 4 float values representing the X coordinate of the 
                object, Y coordinate of the object, the height of the object, Width of the object.
                input: 
                    {}
                Input is always empty as it doesnt require anything as input and analyzes on the image that you are given. Always ignore the arguement priority and do not generate that in the input.

                response:
                    The output is a dict containing object labels as key and a array in XYHW format corresponding the position of the object.

  Tool Args: priority: Optional[str] = "3"
	     	       possible values: ["1","2","3"]
	     model_name: Optional[str] = "DETASwinL"
			 possible values: ["DETASwinL","DETARes","YoloV8s"]

> Tool Name: Optical Character Recognition Tool: ocr\n
  Tool Description: You can use this tool to analyze the given image, The tool should be used when
                you require to extract text from the image. The algorithm returns
                the extracted text which might not be accurate given the limited performance of the OCR model.

                This returns response in a list of strings which is simply in the order of the 
                text present in the image from left to right and top to bottom.
                input: 
                    {}
                Input is always empty as it doesnt require anything as input and analyzes on the image that you are given.
		Always ignore the arguement priority and do not generate that in the input.

                response:
                    The output is a list of string containing the text that is extracted in
		    the order it is present in the image.

  Tool Args: priority: Optional[str] = "3"
	     	       possible values: ["1","2","3"]
	     model_name: Optional[str] = "TROCRLarge"
			 possible values: ["TROCRLarge","TROCRBase","TROCRSmall"]


> Tool Name: Image Recognition Tool: recog

  Tool Description:  You can use this tool to analyze the given image, The tool should be used when
                you require to understand the scene in the image, and get a descriptive text
                about the image. The algorithm returns the description about the image in simple string.

                This returns response in string which is simply contains the description.
                input: 
                    {}
                Input is always empty as it doesnt require anything as input and analyzes on the image that you are given. 
		Always ignore the arguement priority and do not generate that in the input.

                response:
                    The output is a string containing the description.

  Tool Args: priority: Optional[str] = "3"
	     	       possible values: ["1","2","3"]
	     model_name: Optional[str] = "MPLUGLarge"
			 possible values: ["MPLUGLarge","MPLUGBase","BlipT5XXL"]

> Tool Name: Image Recognition Tool: Critic

  Tool Description:  You are supposed to call this tool after you arrived to the answer of the question. 
                This tool will evaluate the answer and provide feedback on the answer.
                input:
                    {}
                    
                    The critic has access to all the information about the React agent and its actions.
                    It also has access to the question and the image for the query.

                Your task is to call it at the end of the reasoning chain and then use the feedback to improve your action and
                solve the query efficiently.
                response:
                    The output is simple text giving feedback and checkboxes based on evaluation criteria.

  Tool Args: None
\end{lstlisting}

\subsubsection*{Input-Output Definition}
\begin{lstlisting}[language=json, linerange={111-135}]
## Output Format
To answer the question, please use the following format.

```
Thought: I need to use a tool to help me answer the question.
Action: tool name (one of vit, object_detect, ocr, recog)
Action Input: the input to the tool, in a JSON format representing the kwargs (e.g. {{"text": "hello world", "num_beams": 5}})
```
Please use a valid JSON format for the action input. Do NOT do this {{'text': 'hello world', 'num_beams': 5}}.

If this format is used, the user will respond in the following format:

```
Observation: tool response
```

You should keep repeating the above format until you have enough information
to answer the question without using any more tools. At that point, you MUST respond
in the following format:

```
Thought: I can answer without using any more tools.
Answer: [your answer here]
```
\end{lstlisting}

\subsubsection*{Guidelines}
\begin{lstlisting}[language=json, linerange={137-144}]
Below is the current conversation consisting of interleaving human and assistant messages.

your task is to solve a given question, this is a vision language task where the question requires to understand the given image/images(if specified in the question). 
            To solve the question you have to take actions in which you can use a tool if required, Vit primarily is used to incorporate in your output using queries this enables you to ask questions about input image/images to an vision expert, this will return rich response containing information from the image/images for your query.

HUMAN: your task is to solve a given question, this is a vision language task where the question requires to understand the given image. To do so you can use the multiple tools to analyze the image, Answer the question: {question} in few words.

\end{lstlisting}

\subsection{Video Pipeline}
\subsection*{Prompt Structure}

The prompt is structured using a series of XML-like tags to clearly delineate different sections. This structure ensures clarity and consistency in presenting the information to the Video Question Answering agent.

\begin{lstlisting}
<tools>
  ... tool definitions ...
</tools>

<guidelines>
  ... guidelines for using tools ...
</guidelines>

<input-output>
  ... input-output format specifications ...
</input-output>
\end{lstlisting}

\subsubsection*{\textless tools\textgreater}

This section defines the available tools for the agent. Each tool is presented with its name, input-output format using Python type hints, and a concise description of its functionality.

\subsubsection*{\textless guidelines\textgreater}

This section provides comprehensive guidelines on how to effectively utilize the tools for answering user questions. It emphasizes strategic tool selection based on the question and the strengths and weaknesses of each tool.

\subsubsection*{\textless input-output\textgreater}

This section meticulously outlines the input-output communication format. It emphasizes the use of clean JSON for all interactions, adhering to standard syntax without any markdown or special characters.

\subsection*{Concrete Prompt Content}

\subsubsection*{\textless tools\textgreater}

\begin{lstlisting}
1)
Tool: get_transcript() -> str:
Description: This tool returns the full transcript of the video along with timestamps for each phrase.

2)
Tool: query_transcript(transcript_query: str) -> str:
Description: This tool allows you to issue a search query over the video transcript and return the timestamps of the top 3 semantically matched phrases in the transcript. The returned timestamps are the average time between the start and end of matched phrases. The timestamps would be comma separated (presented in their matching order with the leftmost being the highest match) and in the format %H:%M:%S (e.g. 00:08:27, 00:23:56, 01:14:39)

3)
Tool: query_frames_Azure_Computer_Vision(frames_query: str) -> str:
Description: This tool allows you to issue a natural language search query over the frames of the video using Azure's Computer Vision API to a find a specific moment in the video. It is good at OCR, object detection and much more. The output format is similar to the query_transcript tool. It returns comma separated timestamps of the top 3 frames that match with given query.

4)
Tool: query_GPT4_Vision(timestamp: -> str, query: -> str) -> str:
Description: This tool is designed to allow you to verify the retrieved timestamps from other tools and also ask more nuanced questions about these localized segments of the video. It utilizes GPT4's Vision capabilities and passes a 10 second clip (only visuals, no audio or transcript) sampled at 1 fps and centered at "timestamp" (which is likely returned by other tools; its format is the same i.e. %H:%M:%S) along with a "query" to the model. Note that this query can be any prompt designed to extract the required information regarding the clip in consideration. The output is simply GPT4's response to the given clip and prompt.
\end{lstlisting}

\subsubsection*{\textless guidelines\textgreater}

\begin{lstlisting}
- For any question, you should always do get_transcript first. This would allow you to directly tackle the questions that are answerable by just looking at the transcript modality. If this is the case, just answer and stop there and do not unnecessarily call other tools. If not, in many cases, the transcript might contain a partial answer, a related event, or any hint/reference indicating where in the visuals the answer might be found. If that is the case then you must diligently note down these details from the transcript in your "observation" and remember them for future use since they will help you in deciding whether to retrieve potentially relevant visuals using query_transcript or not. However, if neither of these are true, then looking at the transcript would still give you a basic understanding of the video and might enable you to answer some generic questions like video summary and also dismissing extremely irrelevant questions. In case the transcript is empty, you must understand that this video only contains visuals and hence focus only on that.
- If the question wasn't fully answerable by the transcript, then it implies that at least some part of the answer lies in the visuals. Now here you must proceed by retrieving potentially relevant timestamps for the visuals and check them one-by-one for relevant information regarding the user query. The checking and reasoning would be done using query_GPT4_Vision but before that you must retrieve the timestamps to feed it in the first place. If the transcript reveals a partial answer or hints/references to a related event corresponding to the user query, the next immediate step is to use query_transcript for retrieving timestamps related to these events or hints. This method should be prioritized as it leverages direct information from the transcript to guide visual analysis. Hence, in this case, start with retrieving timestamps using query_transcript and analyzing them using query_GPT4_Vision and if that is not enough to answer the user_query then you can again retrieve timestamps using query_frames_Azure_Computer_Vision and analyze them using query_GPT4_Vision. On the other hand, if the transcript was empty or had no mention of anything related to the user query whatsoever then directly retrieve timestamps using query_frames_Azure_Computer_Vision and analyze them using query_GPT4_Vision. All of the these steps are clearly explained one-by-one below.
- As mentioned before, if the transcript has a partial answer, a related event, or any hint/reference indicating where in the visuals the answer might be found then you must proceed your visual investigation by trying to retrieve relevant timestamps using query_transcript. Remember that query_transcript allows you to do a semantic search over the transcript by issuing a search query that you will come up with based on the user query/transcript information and it will return the timestamps of the top phrases that match with it where you can analyze the corresponding visuals. On the other hand, if the transcript was empty or had no mention of anything related to the user query whatsoever then you must proceed your visual investigation by trying to retrieve relevant timestamps using query_frames_Azure_Computer_Vision which allows you to issue a visual query (on the frames) that you should come up with based on the user query. Remember that the search query in query_frames_Azure_Computer_Vision is not a prompt; you should think of it as a keyword search that can do OCR, object detection or find some relevant scene based on the given keywords. You should consider all the timestamps returned by these retrievers as potentially important. The first one would be the highest match to the search query and should be explored first.
- Once you have the timestamps from one of these retrievers you should use query_GPT4_Vision. The tool query_GPT4_Vision is a gold standard tool at your disposal. You can give it any relevant timestamp discovered using one of these retrievers and an extensive, nuanced or even open ended prompt about the 10 second clip near that timestamp and it will answer it. You should use this tool to verify and ask more questions about the retrieved timestamps, do any kind of visual reasoning and also to extract final answers from visuals. The idea here is that query_GPT4_Vision can only accept small 10 second clips and hence we do necessary retrieval using query_transcript or query_frames_Azure_Computer_Vision and once we have localized segments we verify and reason using query_GPT4_Vision. Just make sure to not directly refer to these as clip or video in the prompt since GPT4 Vision can only accept still frames. Hence start your prompt with "These are the still frames from a short video clip." and then go on to ask your questions.
- If the transcript had a partial answer or a hint to a related event and you did retrieval using query_transcript but the follow up reasoning using query_GPT4_Vision did not result in satisfactory answers for the user query then you must proceed with follow up retrieval using query_frames_Azure_Computer_Vision and corresponding reasoning using query_GPT4_Vision.
- Remember that you must use these tools to extract information and ground your answer to the user question and not just come up with stuff on your own. If you are unable to properly answer based on the information you initially tried to find then try again. Explore all the different retrievals that you have, change your search queries (to get new retrievals) and keep making logical attempts at exploring the video. If you still unable to answer after trying really hard then you may respond with "I am unable to answer this question" rather than making something up.
- Once you are done with your reasoning and return a final answer you will get feedback from a critic that will carefully analyze your reasoning and answer and let you know if something is not quite right. After you get the feedback, you must continue to methodically reason about the answer while incorporating the critic feedback and the context of your reasoning till that point.
\end{lstlisting}

\subsubsection*{\textless input-output\textgreater}

\begin{lstlisting}
- All communications would be using clean JSON format without any additional characters or formatting. The JSON should strictly follow the standard syntax without any markdown or special characters.
- To start with, you will receive a json with a question.
{
"Question": #some user question
}
- You must respond with a json as follows:
{
"Observation": #observation and comments/understanding of the given question/tool output
"Thought": #plan and think about what should be done next. This can contain both: reasoning about the immediate next step and if needed, also the high level plan about the next few steps
"Action":
{
"tool_name": #select the tool to use based on your observation and thought. E.g. query_GPT4_Vision
"tool_input": 
{
#give the tools inputs as a json with attributes as input names and values as inputs themselves. E.g. {'timestamp':"00:08:27", 'query':"What is happening in this video clip?"}
}
}
}
- You will receive tool outputs using this simple JSON:
{
"Output": #tool output
}
-You will again respond with a json with Observation, Thought and Action (as described before) and this loop will go on N times till you have gathered sufficient information to answer the question.
-Once you think you have enough information to answer, you can replace the "Action" with "Answer" and should respond with the following json:
{
"Observation": #observation and comments/understanding of the given tool output
"Thought": #reasoning on the final answer
"Answer": #answer to user question here
}
-This will then be followed by a critic feedback that will carefully analyze your reasoning and give you feedback on what is missing/wrong. You will receive the critic feedback as follows:
{
"Critic Feedback": #critic's analysis and feedback here
}
-Based on the feedback, you must continue your reasoning:
{
"Observation": #observation and comments/understanding of the given feeback
"Thought": #plan and think about what should be done next. This can contain both: reasoning about the immediate next step and if needed, also the high level plan about the next few steps
"Action":
{
"tool_name": #select the tool to use based on your observation and thought. E.g. query_GPT4_Vision
"tool_input": 
{
#give the tools inputs as a json with attributes as input names and values as inputs themselves. E.g. {'timestamp':"00:08:27", 'query':"What is happening in this video clip?"}
}
}
}
Once you are done, again return the final answer:
{
"Observation": #observation and comments/understanding of the given tool output
"Thought": #reasoning on the final answer
"Answer": #answer to user question here
}
- This will keep happening till the critic is satisfied with your reasoning and answer.
\end{lstlisting}

\section{Tool Augmentation agent: Additional details}
\label{app:tools}

\begin{table}
\centering
\caption{List of Supported Tools and Models by Category}
\label{tab:tools_models}
\begin{tabular}{|c|l|p{6cm}|}
\hline
\textbf{Category} & \textbf{Tools} & \textbf{Models} \\ \hline
\multirow{4}{*}{Image Understanding} & VIT & LLaVA-13B-1.2, InstructBLIP Flan-T5-xxl, InternLM-XComposer2, GPT4V \\ \cline{2-3}
 & OCR & TROCR large, TROCR small, MMOCR \\ \cline{2-3}
 & Object Detection & Deta, SwinL, Deta ResNet, Yolov8s \\ \cline{2-3}
 & Recognition & InstructBLIP, Mplug Base, Mplug Large \\ \hline
Audio Analysis & ASR & Whisper, Azure AI Speech \\ \hline
Textual Analysis & Retrievers & text-embedding-ada-002, text-embedding-3-large \\ \hline
Video Analysis & Video Retriever & Azure Video Retriever \\ \hline
Video Understanding & Multi-modal LLMs & GPT4 Vision \\ \hline
\end{tabular}
\end{table}

\textbf{1. Image Understanding \& Descriptors:} These tools focus on comprehending visual content within a image.\
\textit{(a) VIT (Vision Transformer):} VIT is a state-of-the-art deep learning model specifically designed for image classification and understanding. It breaks down an image into smaller patches, embeds them, and processes them through transformer layers to capture spatial relationships and global context. VIT helps in extracting high-level visual features from images, aiding in tasks such as object recognition, scene understanding, and image captioning. We support multiple models, such as instruct-BLip-flan-xl, InternLM-Composer2, GPT4V. \\
\textit{(b) OCR (Optical Character Recognition):} OCR identifies and extracts text from images, enabling the analysis of textual content within images for tasks such as document analysis, text extraction, and content understanding. We support models such as TROCR large and TROCR small, alongside MMOCR, to ensure robust text recognition across various fonts and backgrounds. \\
\textit{(c) Object Detection:} Object detection identifies and localizes objects within an image. Object detection helps in understanding the visual content of images by identifying and categorizing objects present within them. We support models such as Deta, SwinL, Deta ResNet, and Yolov8s. \\
\textit{(d) Recognition (Face/Object Recognition): }Recognition involves identifying specific objects or faces within images. This tool aids in tasks such as face recognition, object identification, and attribute detection, enhancing the understanding of visual content by recognizing specific entities within images. We support models such as InstructBlip, FlanXL, Mplug Base, and Mplug Large. 

\textbf{2. Audio Analysis \& Descriptors:} We employ Automatic Speech Recognition (ASR) to convert spoken language into text. ASR is crucial for tasks such as transcribing audio data, extracting spoken information, and facilitating multi-modal analysis by incorporating audio-based information into the overall understanding. We support models like Whisper, Azure AI Speech, etc.

\textbf{3. Textual Analysis \& Retrievers:}
This tool identifies and returns the timestamps of the top semantically matched phrases in the transcript given a search query. Retrieving text from transcripts helps in tasks such as information retrieval, context understanding, and text-based analysis in multi-modal scenarios. We employ embedding models like text-embedding-ada-002 and text-embedding-3-large from OpenAI to encode each phrase from the transcript and the user’s search query and return top matches using cosine similarity. 

\textbf{4. Video Analysis \& Retrievers:}
This tool analyses the video frames (either all frames or sub-sampled) and create a queriable index. To accomplish this, video indexer tools use video embeddings like CLIP, etc., to create the indexes that allow for the identification of specific moments through natural language search queries. Video Retriever aids in tasks such as video summarization, content analysis, and object tracking, enhancing the understanding of visual information within videos. We support Azure Video Retriever from Microsoft Azure. The tool returns the top-3 frames that best match the given query.

\textbf{5. Video Understanding \& Descriptors:} This tool provides foundational models that can thoroughly analyse multiple video frames simultaneously. This allows the agent to understand and reason over multi-modal data by leveraging both textual and visual cues for comprehensive analysis and decision-making. We provide support for GPT4 Vision as part of the visual understanding toolset. The tool processes a 10-second clip centered around the provided timestamp of interest, sampled at one frame per second (fps) and passes that along with the given prompt to the selected MLLM. 

Each of these tools plays a vital role in augmenting \mmct's capabilities to comprehend multi-modal information by extracting relevant features, recognizing entities, and facilitating analysis across different modalities, ultimately enhancing the overall understanding and reasoning process.

\section{Vision-based Critic: Additional details}
\label{app:critic}
\subsection{Image Pipeline}
In this section, we describe the prompts and the structure of the Criteria utilized in our pipeline. Along with the prompts used to generate Criteria we also discuss the prompt of the critic. 
\textbf{Criteria Generation}: To generate the criteria, we use a prompt that can be decomposed into \texttt{[Problem description]}, \texttt{[Instruction]}, \texttt{[Task Description]} and \texttt{[Human Intent]}. Each part is explained in \ref{sec:critic}; here, we present the prompt and the specific inputs used for our pipeline.

\subsection*{Problem description}
\begin{lstlisting}[language=json, linerange={1-7}]
Criteria Prompt:
 
You are a helpful assistant. You suggest criteria for evaluating different tasks.  
Define the evaluation criteria as a dictionary where the keys are the criteria. 
The value of each key is a dictionary which includes a description of the criteria and evaluation value. 
Include the evalaution value as fine-grained and multi level.
\end{lstlisting}
\subsection*{Instruction}
\begin{lstlisting}[language=json, linerange={7-15}]

Make sure the keys are criteria for assessing the given task. 
evaluation values: include the acceptable values for each key that are fine-grained and preferably multi-graded levels.

You are given the task description as Task Description: {TaskDescription}. 
Further, you are given few human intents that should be taken into account for defining the criteria. 
Human Intent: {HumanIntent}
        
Given these information, you have to give me a list of 5 criterions such that one can inspect to ensure the task is solved without any discrepancies.
\end{lstlisting}
\subsection*{Task Description}
\begin{lstlisting}[language=json, linerange={17-25}]
TaskDescription:

Visual Question Answering:
        An Image is given along with a query which needs to be addressed,
        This task requires a great amount of reasoning to achieve the task,
        while also requiring varied capabilities based on the query like OCR, Spatial
        Understanding, Common Knowledge, etc. The objective is to give a satisfactory
        answer to the query.
\end{lstlisting}
\subsection*{Human Intent}
\begin{lstlisting}[language=json, linerange={26-28}]
HumanIntent:
        concise answer, clear reasoning, comprehensive, uses all necessary tools for answering
\end{lstlisting}

Using the above prompt, a sample output of Criteria is as seen below
\subsection*{Sample Criteria}
\begin{lstlisting}[language=json, linerange={29-50}]
Criteria:

1. Accuracy of Answer:
    - Description: The correctness of the answer provided by the model in response to the query.
    - Evaluation Values: {"1": Incorrect, "2": Partially Correct, "3": Mostly Correct, "4": Almost Correct, "5": Correct}

2. Conciseness:
    - Description: The ability of the model to provide a concise and to-the-point answer without unnecessary information.
    - Evaluation Values: {"1": Very Verbose, "2": Somewhat Verbose, "3": Moderately Concise, "4": Mostly Concise, "5": Highly Concise}

3. Clarity of Reasoning:
    - Description: The clarity and logical flow of the model's reasoning process in arriving at the answer.
    - Evaluation Values: {"1": No Reasoning, "2": Poor Reasoning, "3": Adequate Reasoning, "4": Good Reasoning, "5": Excellent Reasoning}

4. Comprehensiveness:
    - Description: The extent to which the model considers all relevant aspects of the image and query to provide a well-rounded answer.
    - Evaluation Values: {"1": Not Comprehensive, "2": Slightly Comprehensive, "3": Moderately Comprehensive, "4": Mostly Comprehensive, "5": Fully Comprehensive}

5. Use of Necessary Tools:
    - Description: The model's ability to utilize appropriate tools and techniques (e.g., OCR, spatial understanding, common knowledge) to answer the query effectively.
    - Evaluation Values: {"1": No Tools Used, "2": Few Tools Used, "3": Some Tools Used, "4": Most Tools Used, "5": All Necessary Tools Used}
\end{lstlisting}

Here, for each criterion, the evaluation value is a dictionary defining a grade, with \texttt{"1"} being the lowest acceptable grade and \texttt{"5"} being the best grade that a critic can assign to a comprehensive reasoning chain.

With the obtained Criteria, we input it in the Critic prompt, as seen below, to obtain the assessed evaluation values and feedback to iterate the chain and continue the cycle until all the evaluation values meet an acceptable range.

\subsection{Video Pipeline}
\subsection*{Prompt Structure}

The critic prompt is structured similarly to the agent prompt, using XML-like tags to delineate different sections. This structure ensures clarity and consistency in presenting the information to the critic.

\begin{lstlisting}
<tools>
... tool definitions ...
</tools>

<critic_guidelines>
... guidelines for evaluating agent reasoning ...
</critic_guidelines>

<input-output>
... input-output format specifications ...
</input-output>

<sample_response>
... sample response in JSON format ...
</sample_response>
\end{lstlisting}

\subsubsection*{\textless tools\textgreater}

This section defines the same set of tools available to the agent. This ensures the critic understands the capabilities and limitations of the tools used in the reasoning chain.

\subsubsection*{\textless critic\_guidelines\textgreater}

This section provides comprehensive guidelines for the critic to evaluate the agent's reasoning. It outlines three key criteria:

\textbf{Answer Completeness}: Assess whether the user query is fully answered, partially answered, or not answered at all.

\textbf{Reasoning Comprehensiveness}: Analyze the thoroughness of the reasoning chain, ensuring the agent explored all relevant avenues and utilized the tools effectively.

\textbf{Hallucination Detection}: Identify any instances where the agent might have generated information not grounded in the provided video data, either through misinterpreting tool outputs or fabricating answers.

\subsubsection*{\textless input-output\textgreater}

This section meticulously outlines the input-output communication format for the critic. The critic's response should include:

\textbf{Observation}: A detailed analysis of the agent's logs based on the critic guidelines.

\textbf{Thought}: The critic's assessment of the reasoning chain's correctness based on the observation and criteria.

\textbf{Feedback}: Specific feedback for each criterion, highlighting any issues and offering suggestions for improvement.

\textbf{Verdict}: A final "YES" or "NO" verdict on the overall correctness of the reasoning chain.

\subsubsection*{\textless sample\_response\textgreater}

This section provides a concrete example of a correctly formatted JSON response from the critic, including placeholder strings for each key. This serves as a template for the critic to follow when providing feedback.

\subsection*{Concrete Critic Prompt Content}

\subsubsection*{\textless tools\textgreater}

\begin{lstlisting}
1)
Tool: get_transcript() -> str:
Description: This tool returns the full transcript of the video along with timestamps for each phrase.

2)
Tool: query_transcript(transcript_query: str) -> str:
Description: This tool allows the reasoning agent to issue a search query over the video transcript and return the timestamps of the top 3 semantically matched phrases in the transcript.

3)
Tool: query_frames_Azure_Computer_Vision(frames_query: str) -> str:
Description: This tool allows the reasoning agent to issue a natural language search query over the frames of the video using Azure's Computer Vision API to find a specific moment in the video. It is good at OCR, object detection, and much more.

4)
Tool: query_GPT4_Vision(timestamp: -> str, query: -> str) -> str:
Description: This tool is designed to allow the reasoning agent to verify the retrieved timestamps from other tools and also ask more nuanced questions about these localized segments of the video. It utilizes GPT4's Vision capabilities and passes a 10 second clip (only visuals, no audio or transcript) sampled at 1 fps and centered at "timestamp" along with a "query" to the model. Note that this query can be any prompt designed to extract the required information regarding the clip in consideration. The output is simply GPT4's response to the given clip and prompt.
\end{lstlisting}

\subsubsection*{\textless critic\_guidelines\textgreater}

\begin{lstlisting}

Analyse whether the user query is fully answered, partially answered, or not answered.

Analyse the comprehensiveness of the reasoning chain in the sense that whether thorough analysis was done; for example, whether query_transcript was used to find relevant timestamps for answering the question if the transcript returned by get_transcript had something related to the question or whether the system tried hard to find the answer before giving up in the case that it couldn't answer etc.

Analyse whether there are any hallucinations in the sense that whether the query_GPT4_Vision calls actually returned info true to the images given to you or did it return something from its general knowledge; whether the reasoning chain returned the final answer based on its analysis or hallucinated it etc.
\end{lstlisting}

\subsubsection*{\textless input-output\textgreater}

\begin{lstlisting}

All communications would be using clean JSON format without any additional characters or formatting. The JSON should strictly follow the standard syntax without any markdown or special characters.

To start with, you will receive a json with the logs.
{
"logs": #some agent logs
}

For your response, you must proceed as follows:
{
"Observation": #observation and analysis of the given logs by taking into account all the critic guidelines
"Thought": #think about whether the logs were correct or wrong based on the observation and criteria
"Feedback":
{
"Criteria 1": #craft careful feedback based on your analysis and the first criteria in critic guidelines; if its fine then just declare that otherwise point out what is wrong and if possible also give some suggestions on what the agent might do next; for example you might suggest it to retrieve and analyse additional timestamps using some particular search query to complete a partially answered question
"Criteria 2": #craft careful feedback based on your analysis and the second criteria in critic guidelines; if its fine then just declare that otherwise point out what is wrong and if possible also give some suggestions on what the agent might do next; for example if the agent overlooked some detail in the question you might suggest it to use query_GPT4_Vision with a slightly different query for correctness or retrieve timestamps using some different search query etc
"Criteria 3": #craft careful feedback based on your analysis and the third criteria in critic guidelines; if its fine then just declare that otherwise point out what is wrong and if possible also give some suggestions on what the agent might do next; for example if you think a particular timestamp was hallucinated then ask the agent to check that again with query_GPT4_Vision
}
"Verdict": #Based on the Feedback, come up with a final "YES" or "NO" verdict on whether the reasoning was fine or not; "YES" means completely fine and "NO" means not fine i.e. at least one of the criteria was not perfectly satisfied; only return "YES" or "NO"
}
\end{lstlisting}

\subsubsection*{\textless sample\_response\textgreater}

\begin{lstlisting}
{
"Observation": "This is a placeholder observation string.",
"Thought": "This is a placeholder thought string.",
"Feedback": {
"Criteria 1": "This is a placeholder string for Criteria 1 feedback.",
"Criteria 2": "This is a placeholder string for Criteria 2 feedback.",
"Criteria 3": "This is a placeholder string for Criteria 3 feedback."
},
"Verdict": "This is a placeholder verdict string."
}
\end{lstlisting}

\subsection*{Implementation Details for Frames Handling}
\label{app:photogrid}

In our video question answering system, the critic component requires a thorough examination of frames from video segments where our agent conducted analyses. This is crucial for verifying the accuracy and relevance of the information retrieved by the agent.

\textbf{Challenges with Frame Processing:}
Our system faces a technical constraint due to the Azure OpenAI API, which limits the number of frames that can be processed in a single GPT4 Vision API call to 10 frames. This issue is that each GPT4 Vision call by the agent itself uses 10 frames sampled at 1 fps around the queried timestamp. To address this, we devised a method to efficiently distribute these frames across multiple timestamps into these 10 available images for the critic call.

\textbf{Frame Distribution Strategy:}
In the event of multiple GPT4 Vision calls during a reasoning chain, our approach must efficiently manage these frame sets. We prioritize the last 10 timestamps if there are more than 10 in a single sequence. Now consider a specific scenario in MMCT-QA where the agent makes three such calls at timestamps 00:00:36, 00:02:13, and 00:01:23. To adhere to the API's limitations, we distribute these timestamps within the available 10 images. This distribution allows for consistent examination and avoids missing potential visual data.

We distribute them as follows:
\begin{itemize}
    \item Image(s) 1, 2, 3 are for timestamp 00:00:36.
    \item Image(s) 4, 5, 6 are for timestamp 00:02:13.
    \item Image(s) 7, 8, 9, 10 are for timestamp 00:01:23.
\end{itemize}

Further, within a specific timestamp, such as 00:00:36, we distribute the 10 frames among the available images by stacking the frames horizontally. This might be done as:
\begin{itemize}
    \item Image 1 contains 3 frames.
    \item Image 2 contains 3 frames.
    \item Image 3 contains 4 frames.
\end{itemize}

This allocation ensures that each frame is utilized optimally, providing comprehensive visual data for the critic's analysis.

\textbf{Visual Examples:}
Figure~\ref{appfig:critic_grid} illustrates the images corresponding to the timestamp 00:00:36 with their frames distributed as described:

\begin{figure}
\centering
\begin{subfigure}{\textwidth}
  \centering
  \includegraphics[width=0.95\linewidth]{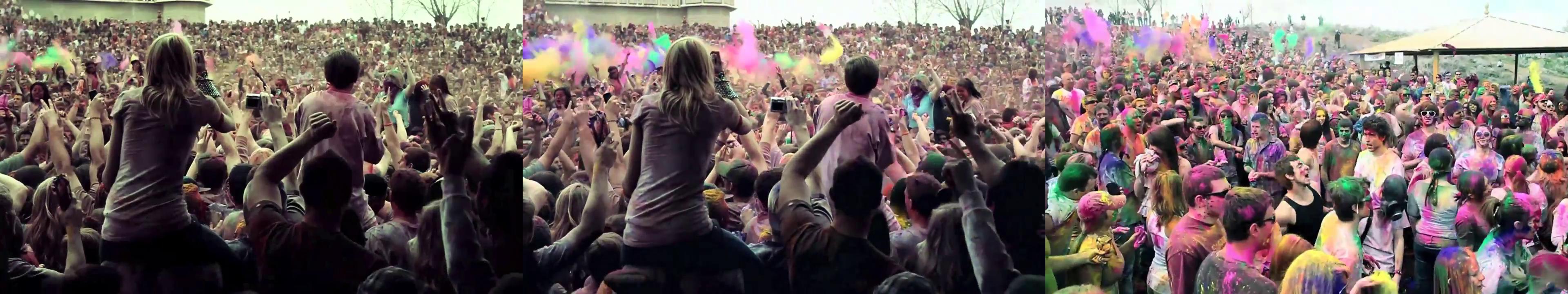}
  \caption{Image 1 with 3 frames}
\end{subfigure}\\[1em]
\begin{subfigure}{\textwidth}
  \centering
  \includegraphics[width=0.95\linewidth]{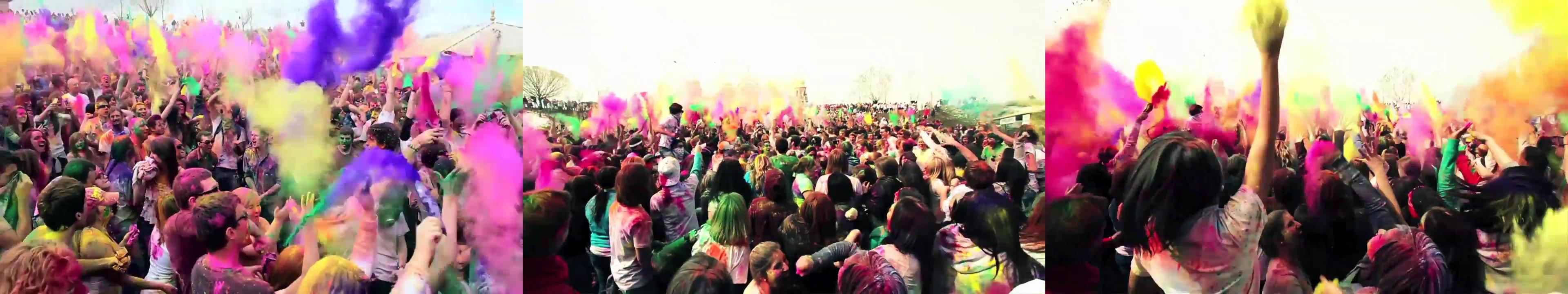}
  \caption{Image 2 with 3 frames}
\end{subfigure}\\[1em]
\begin{subfigure}{\textwidth}
  \centering
  \includegraphics[width=0.95\linewidth]{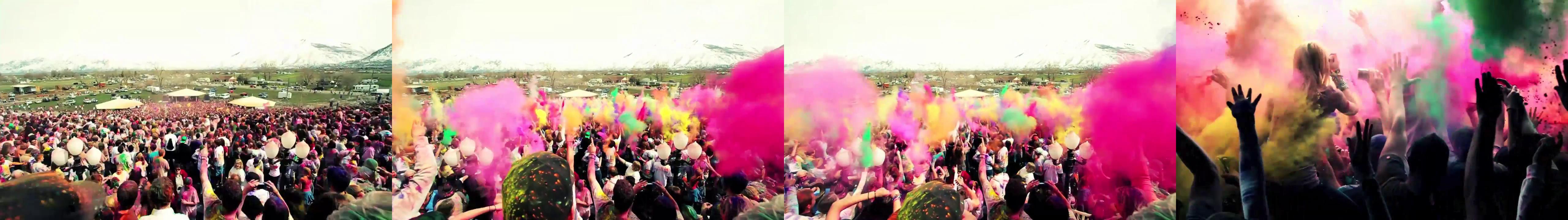}
  \caption{Image 3 with 4 frames}
\end{subfigure}
\caption{Frame distribution for timestamp 00:00:36}
\label{appfig:critic_grid}
\end{figure}

This structured approach ensures that the critic has access to all necessary visual information, aiding in accurate and comprehensive analysis of the video question answering system’s performance.

\section{Image understanding and reasoning: Qualitative examples}
\label{app:image_quali}

\begin{figure}[!t]
\centering
    \includegraphics[trim=0cm 4cm 0cm 0cm, clip,width=0.5\columnwidth]{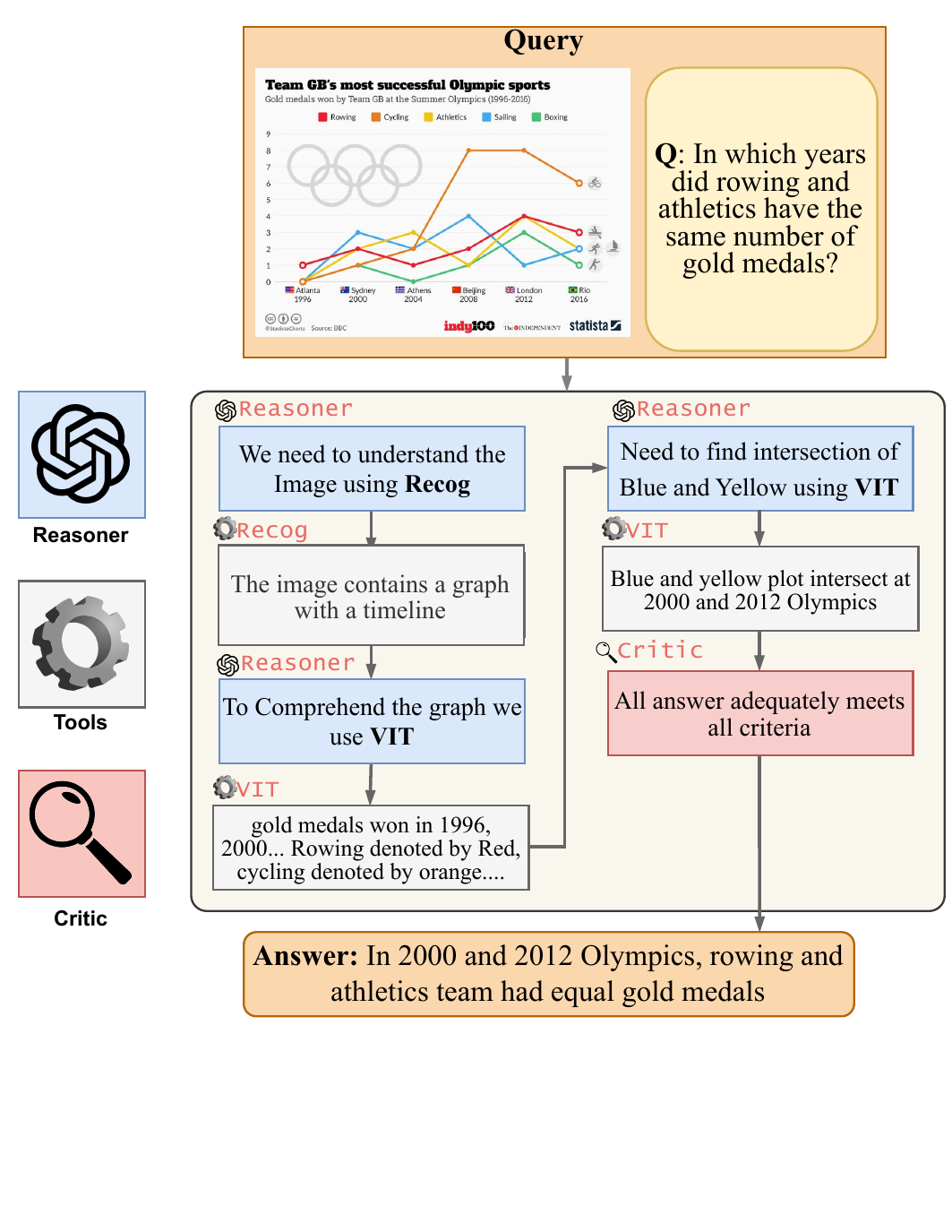}
    \caption{Illustration of a Image QA Qualitative Example with \mmct.}
    \label{appfig:image_qualitative}
    \vspace{-5pt}
\end{figure}

Figure~\ref{appfig:image_qualitative} provides an example to illustrate \mmct`s full execution flow for an image from MMVET dataset. For the given image, the user query is " In which years did rowing and athletics have the same number of gold medals?". This is an example of complex visual reasoning task, where one has to first understand the context of the image, identify different plots, estimate their values and then determine the the instances when they are equal. This specific image resulted in wrong answer with all the foundational MLLMs like GPT4V, Gemini, etc. Let us now see how \mmct~solves this by applying the iterative reasoning process. The planner utilizes \texttt{Recog Tool} to identify the contents of the images proceeding with \texttt{VIT tool} to derive information about the graph and what they denote in the graph. With this information the planner and the reasoning agent proposes to find the intersection of the two plots over individually estimating the value of gold medals one for all team every year. This saved a lot of compute and also giving the right answer. Finally \texttt{VIT tool} is used to answer the years when the blue and yellow plots intersect and it finally gives the answer of 2000 and 2012 olympics. We then invoke the critic agent with the same criteria to evaluate the final answer, the critic also agrees and suggests that the final answer is accurate with coherent reasoning chain.

\section{Video understanding and reasoning: Qualitative examples}
\label{app:video_quali}
\begin{figure}[!t]
\centering
    \includegraphics[width=0.5\columnwidth]{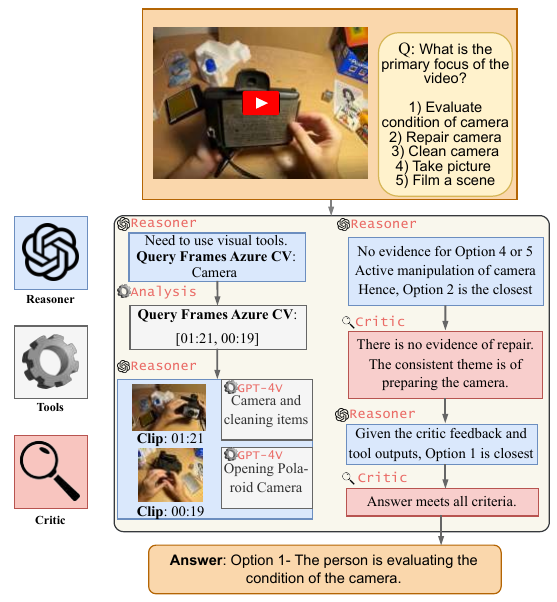}
    \caption{Illustration of a Video QA Qualitative Example with \mmct}
    \label{appfig:video_qualitative}
    \vspace{-5pt}
\end{figure}

Figure~\ref{appfig:video_qualitative} illustrates the \mmct's approach to determining the primary focus of a video involving various potential interactions with a camera by a person (C). The query presented multiple choice answers, posing a unique challenge that involves discerning between evaluating, repairing, cleaning, taking pictures, or filming with the camera. This task necessitates robust visual and critical analysis to interpret the actions of C accurately.

\textbf{Initial Assessment:} The reasoning agent begins by acknowledging the need for visual analysis tools to identify the actions performed by C with the camera, as the query demands identification of the primary focus based on visible interactions (step 1).

\textbf{Visual Querying:} Employing the query\_frames\_Azure\_Computer\_Vision tool, the agent retrieves key frames where C interacts with the camera. The timestamps identified are 00:01:21 and 00:00:19, suggesting these moments are crucial for analysis (step 2).

\textbf{Visual Analysis:} The agent uses the query\_GPT4\_Vision tool to analyze the frames at these timestamps. At 00:01:21, the tool observes C examining and possibly preparing a Polaroid camera, interacting with various parts of the camera in a meticulous manner. Similarly, at 00:00:19, C is noted to be loading film into the camera, further suggesting preparation activities rather than usage (step 3).

\textbf{Critic Evaluation:} Initially, the reasoning led to a hypothesis that C might be repairing the camera due to the active manipulation observed. However, the critic agent points out that there is no evidence of repair or damage; instead, the activities align more with evaluating or preparing the camera. The consistent presence of items related to photography supports a scenario of preparation rather than repair (step 4).

\textbf{Revised Analysis:} Taking into account the critic's feedback and reevaluating the visual evidence, the agent concludes that the primary focus of C is evaluating the condition of the camera, which includes meticulous handling and setup activities, rather than filming or taking immediate pictures with it (step 5).

\textbf{Final Decision:} Integrating insights from both the visual analysis and critic feedback, the agent selects Option 1: "C is evaluating the condition of the camera" as the answer. This conclusion is based on the detailed observations of C's interactions with the camera, focusing on examination and preparation, which are indicative of an evaluation process (step 6).

This reasoning chain successfully demonstrates the \mmct's capability to parse complex visual data and interpret nuanced user queries effectively.

\section{Image Understanding Benchmark Datasets}
\label{app:image_data}
Below we provide details on the five datasets we employ ofr evaluating \mmct. 

\underline{\texttt{MMVET}} dataset~\cite{yu2023mmvet} Evaluates large multimodal models on integrated capabilities across recognition, knowledge, OCR, spatial awareness, language generation, and math, using 200 images and 218 questions to reflect realistic scenarios.

\underline{\texttt{A12D}} dataset~\cite{Kembhavi2016ADI} comprises over 5,000 science diagrams and corresponding questions, testing models' ability to interpret complex visual data crucial for educational and scientific contexts.

\underline{\texttt{MMMU}} dataset~\cite{yue2023mmmu} features 11.5K questions spanning six disciplines, demanding models to apply domain-specific knowledge and reasoning skills across diverse subject matter, from humanities to engineering.

\underline{\texttt{mmbench}} dataset~\cite{liu2024mmbench} consists of around 3,000 questions across 20 ability dimensions, offering a comprehensive evaluation of models' perceptual and reasoning capabilities across various cognitive tasks.

\underline{\texttt{A-OKVQA}} dataset~\cite{marino2019okvqa} presents over 14,000 questions challenging models to integrate external knowledge beyond visual and textual data, reflecting real-world scenarios where broader information is necessary for accurate responses.

\underline{\texttt{MathVista}}, dataset~\cite{lu2024mathvista} is a benchmark designed to evaluate mathematical reasoning in visual contexts. It includes 6,141 examples from 28 existing multimodal datasets and three new datasets: IQTest, FunctionQA, and PaperQA. These datasets focus on algebraic, arithmetic, geometric, logical, numeric commonsense, scientific, and statistical reasoning, covering tasks such as figure question answering, geometry problem solving, math word problems, textbook question answering, and visual question answering.

\section{MMCT-QA Dataset: Details}
\label{app:mmct_data}
Recognizing the limitations inherent in current video question answering datasets, we observed a distinct lack of representation for long-form videos, which not only utilize both audio and visual modalities but also encompass a diverse array of question types beyond specific tasks such as activity recognition. To bridge this gap, we devised a taxonomy of queries classified into six distinct categories, each designed to test different aspects of the system's video understanding capabilities (Table~\ref{taxonomy}).

Building on the framework established by our taxonomy, we proceeded to construct a dataset tailored to test each query category effectively. We selected a subset of 15 diverse videos from the Youtube 8M dataset~\cite{abuelhaija2016youtube8m}, ensuring a variety of content that encompasses different scenarios and interactions. To facilitate the generation of questions and answers, we divided these videos among three human annotators, assigning five videos to each. Each annotator was provided with the taxonomy categories, detailed descriptions, and illustrative examples of generic questions per category to guide their query formulation. This structured approach allowed the annotators to craft questions that are not only relevant to the videos but also representative of each category's specific challenge. As a result of this process, we curated a total of 129 questions, distributed nearly evenly across the six categories (see Table \ref{accuracy}), thereby enabling a comprehensive evaluation of the video question  answering system's capabilities. 
We have presented examples of questions from each category below. 

\begin{table}[htbp]
\caption{Taxonomy for Video Question Answering}
\label{taxonomy}
\centering
\begin{tabular}{p{0.35\linewidth}p{0.58\linewidth}}
\toprule
\textbf{Query Category} & \textbf{Description} \\
\midrule
Temporal Understanding & Assessing system's grasp of event sequences and timing. \\
Spatial Understanding & Evaluating the understanding of spatial relationships and settings within the video. \\
Event and Action Recognition & Focuses on specific actions or events in the video. \\
Dialogue and Transcript-Based & Relying on interpretation of spoken words and its context. \\
Abstract and Conceptual & 
Ability to grasp abstract concepts or themes. \\
Specific Detail Based & Targeted at extracting precise information or details. \\
\bottomrule
\end{tabular}
\end{table}


\begin{table}
\caption{Category-Wise Accuracy of \mmct on MMCTQA}
\label{accuracy}
\centering
\begin{tabular}{p{0.35\linewidth}p{0.04\linewidth}p{0.15\linewidth}}
\toprule
\textbf{Category} & \textbf{\#Q} & \textbf{Accuracy (\%)} \\
\midrule
Temporal Understanding & 22 & 52.3 \\
Spatial Understanding & 22 & 70.5 \\
Event \& Action Recognition & 25 & 62.0 \\
Dialogue \& Transcript-Based & 14 & 89.3 \\
Abstract and Conceptual & 23 & 91.3 \\
Specific Detail Based & 23 & 69.6 \\
\bottomrule
\end{tabular}
\end{table}

\begin{figure}[htbp]
\centering
\includegraphics[width=0.09\textwidth, height=0.1\textwidth]{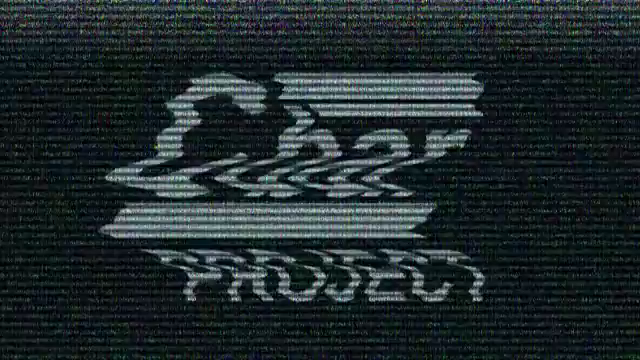}
\includegraphics[width=0.09\textwidth, height=0.1\textwidth]{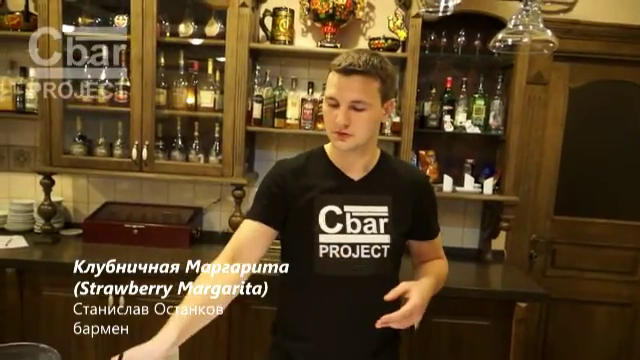}
\includegraphics[width=0.09\textwidth, height=0.1\textwidth]{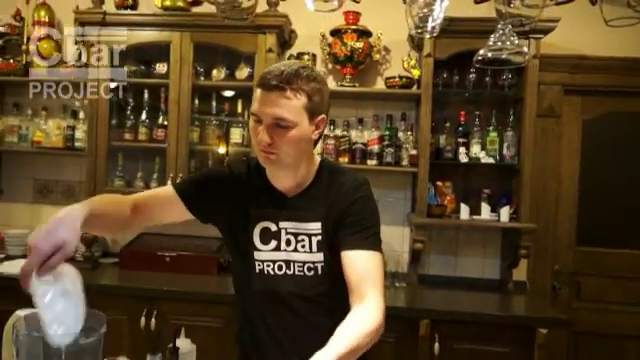}
\includegraphics[width=0.09\textwidth, height=0.1\textwidth]{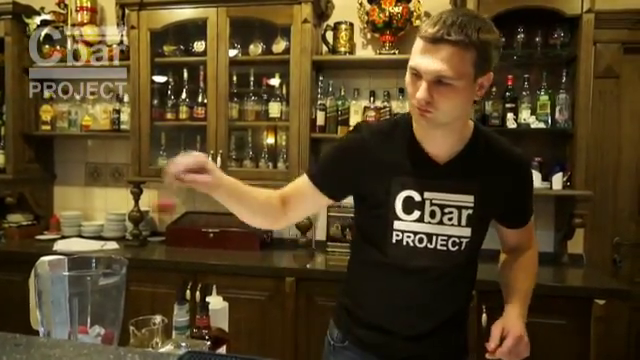}
\includegraphics[width=0.09\textwidth, height=0.1\textwidth]{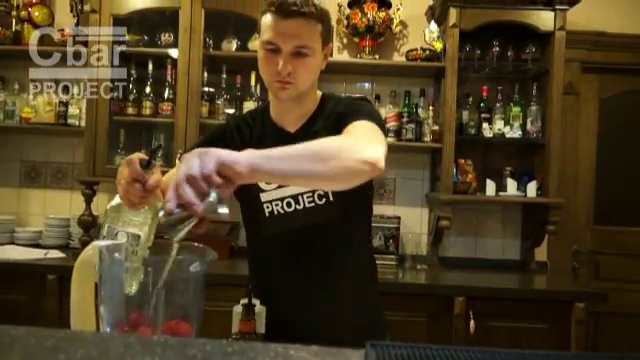}
\includegraphics[width=0.09\textwidth, height=0.1\textwidth]{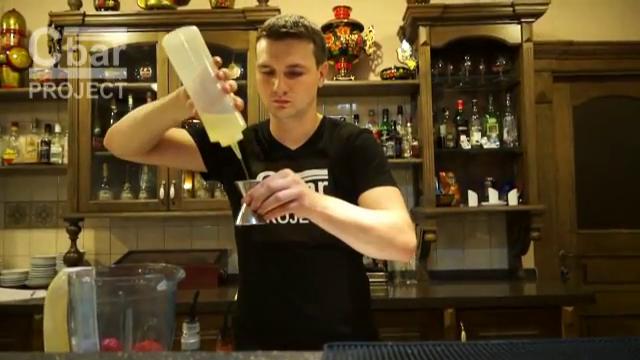}
\includegraphics[width=0.09\textwidth, height=0.1\textwidth]{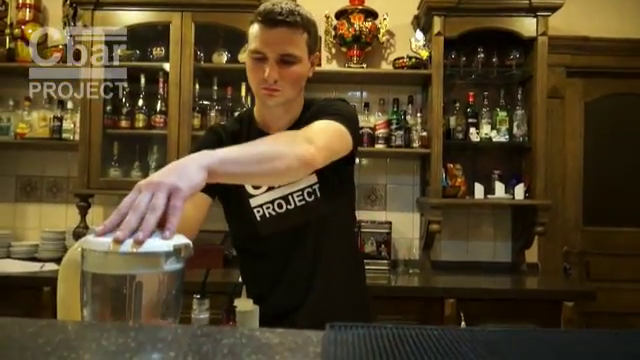}
\includegraphics[width=0.09\textwidth, height=0.1\textwidth]{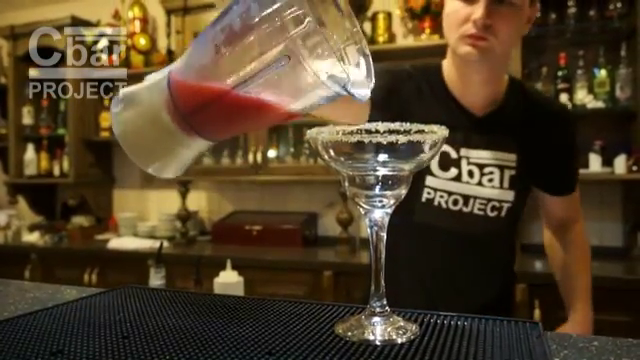}
\includegraphics[width=0.09\textwidth, height=0.1\textwidth]{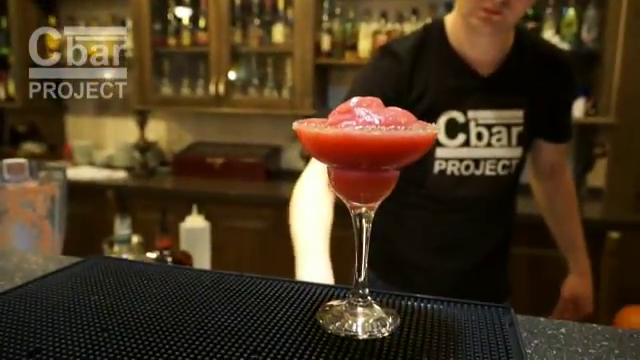}
\includegraphics[width=0.09\textwidth, height=0.1\textwidth]{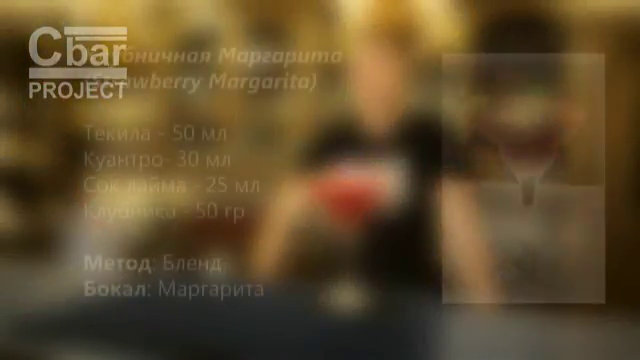}

\medskip 

\textbf{Question:} What is the sequence of things the person added in the mixer?

\medskip 

\textbf{Answer:} The person adds ice, strawberries, tequila, cointreau, \& lime juice to the mixer in that order.

\caption{Example of a Temporal Understanding Question.}
\end{figure}

\begin{figure}[htbp]
\centering
\includegraphics[width=0.09\textwidth, height=0.1\textwidth]{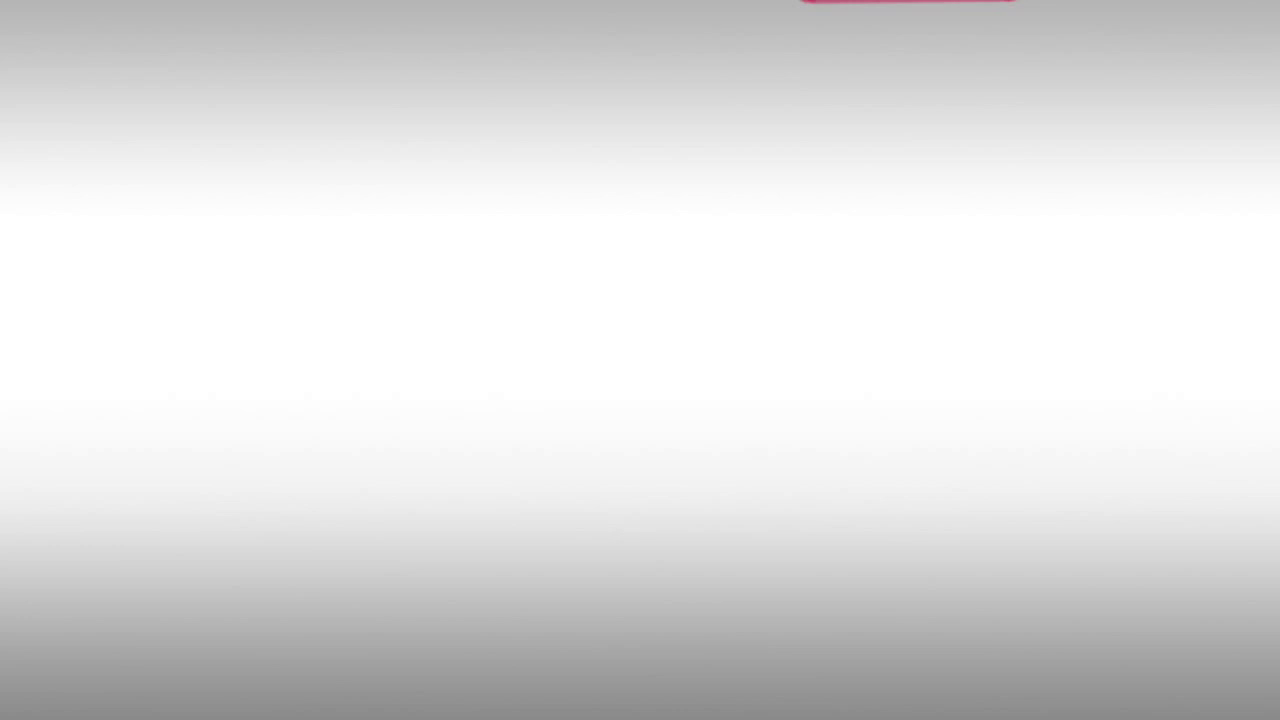}
\includegraphics[width=0.09\textwidth, height=0.1\textwidth]{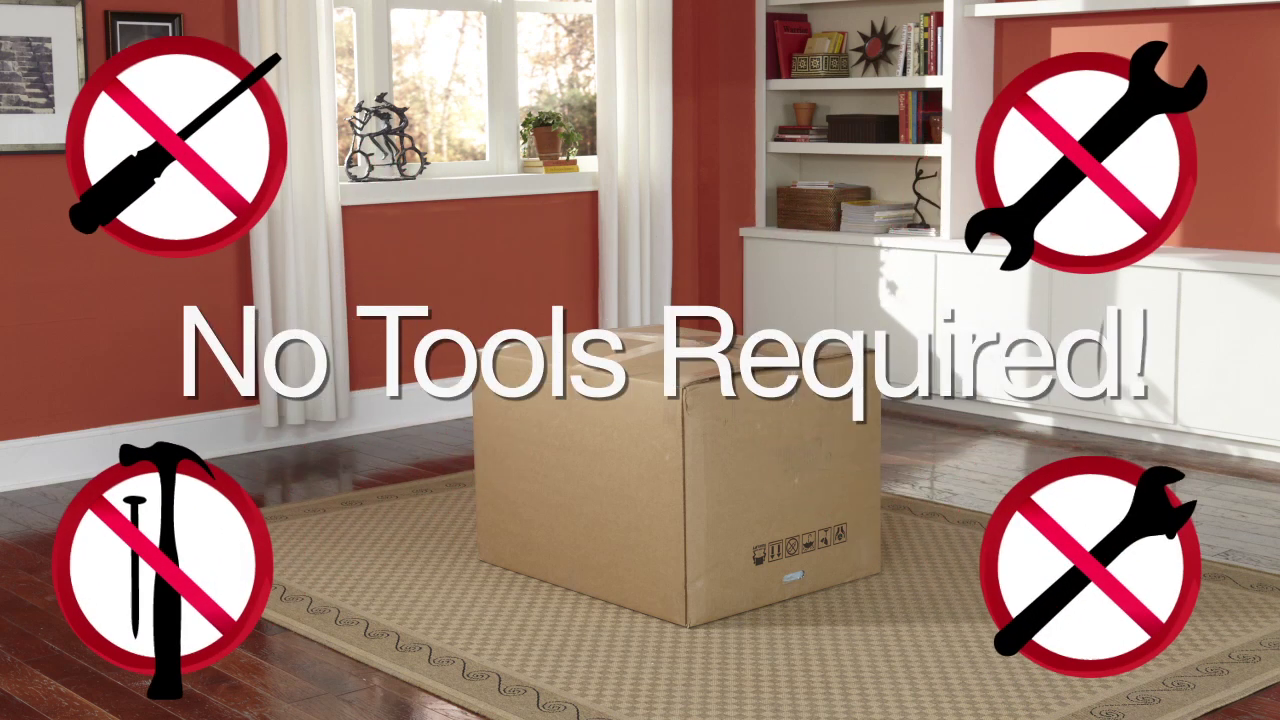}
\includegraphics[width=0.09\textwidth, height=0.1\textwidth]{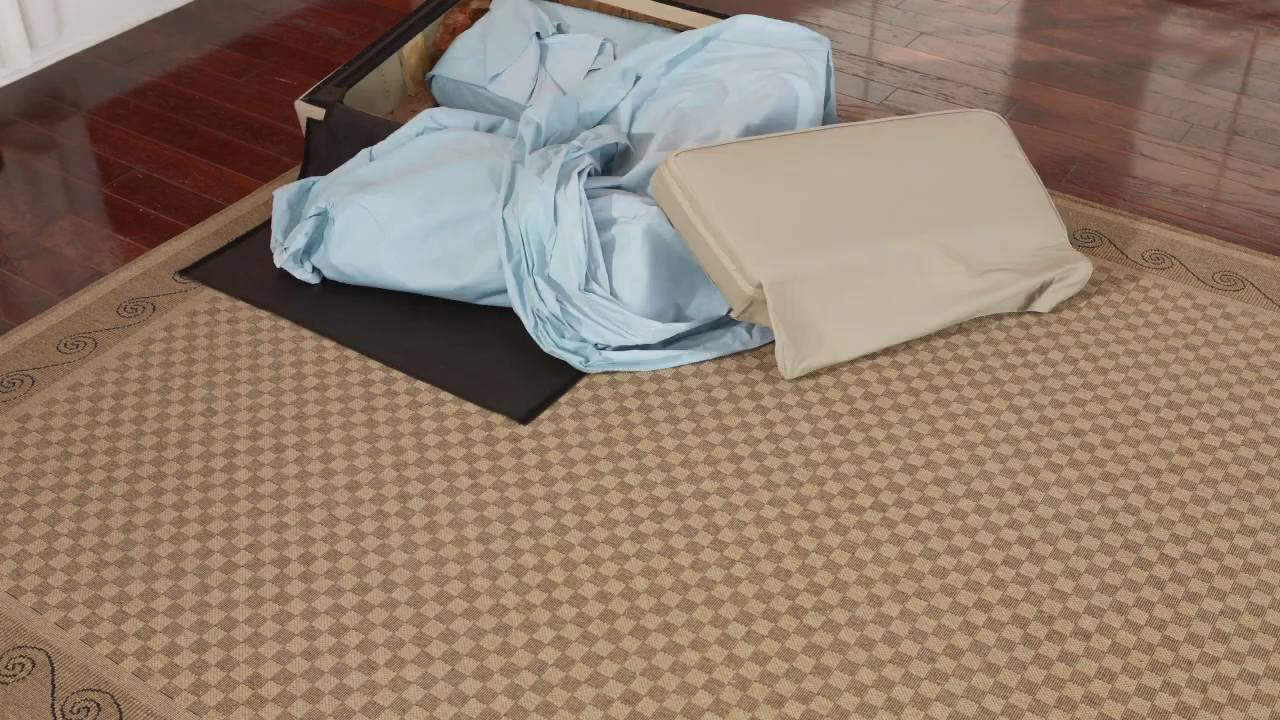}
\includegraphics[width=0.09\textwidth, height=0.1\textwidth]{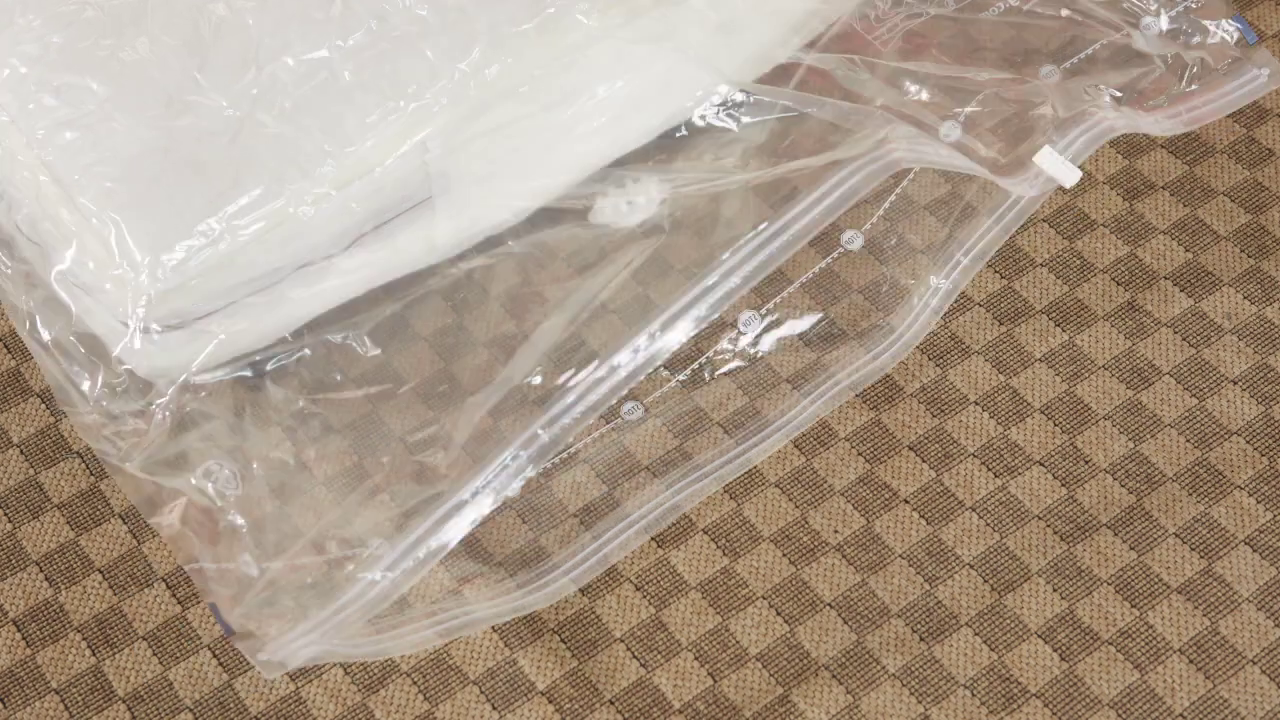}
\includegraphics[width=0.09\textwidth, height=0.1\textwidth]{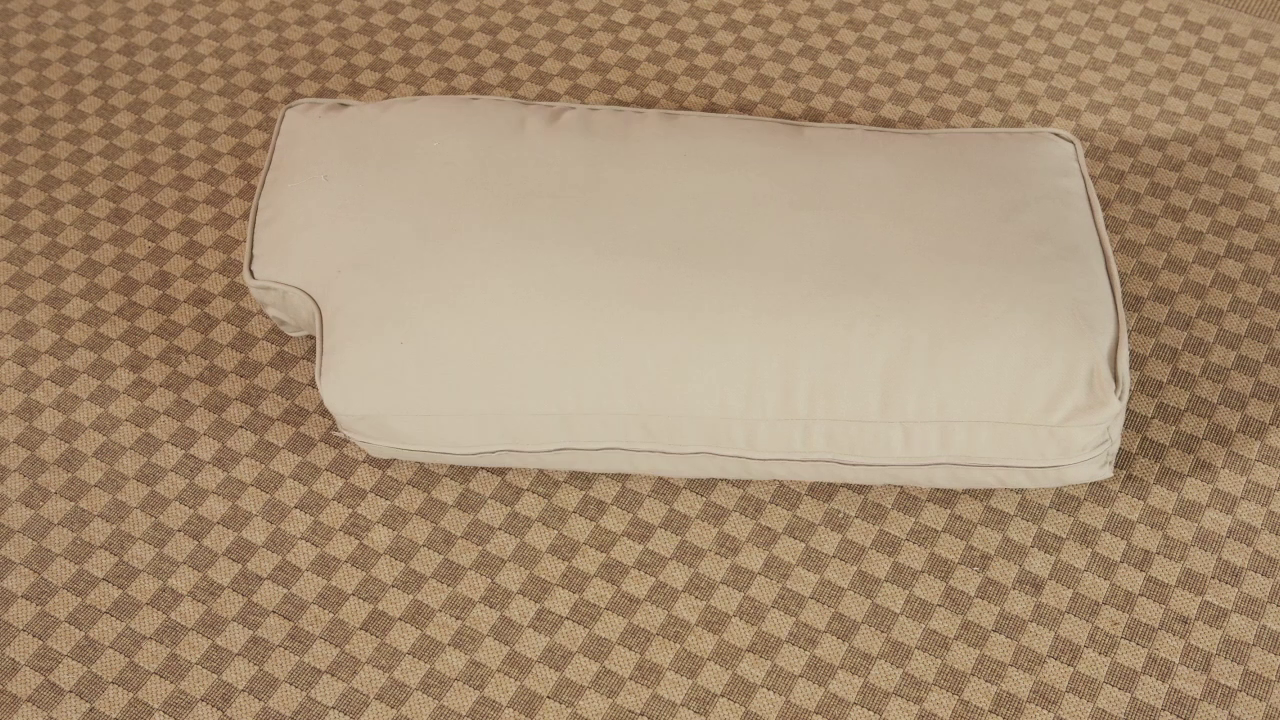}
\includegraphics[width=0.09\textwidth, height=0.1\textwidth]{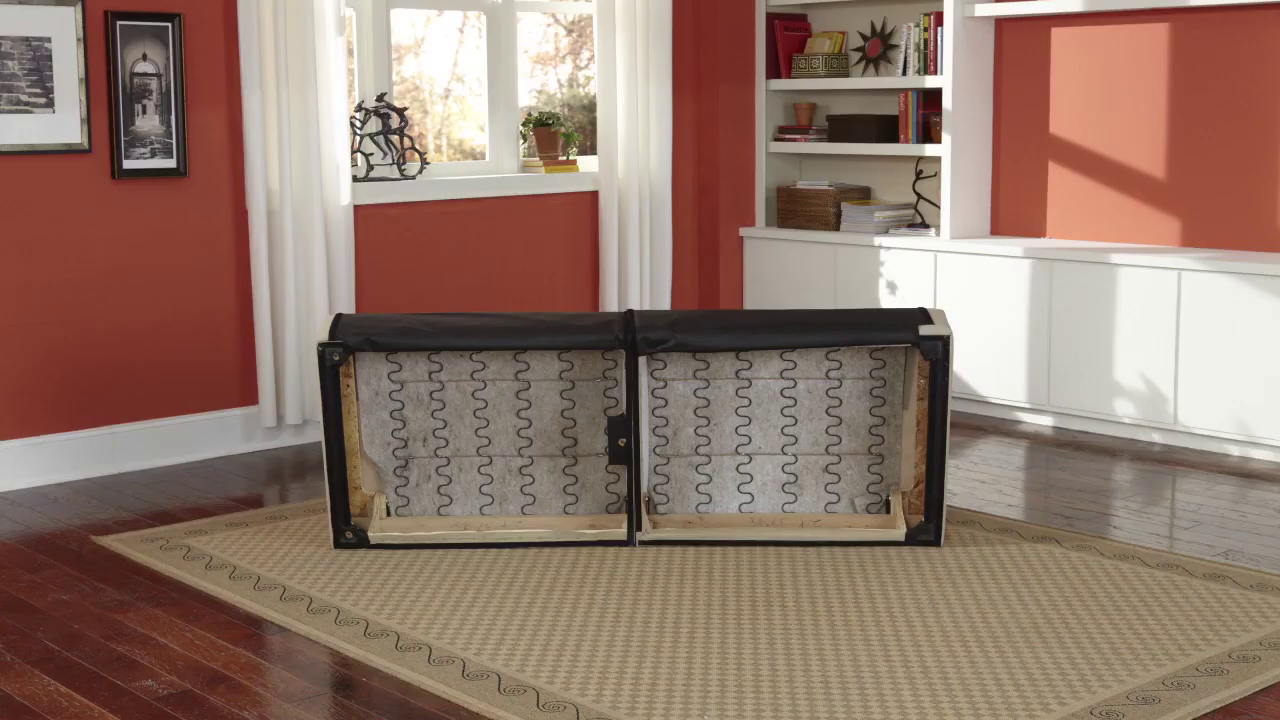}
\includegraphics[width=0.09\textwidth, height=0.1\textwidth]{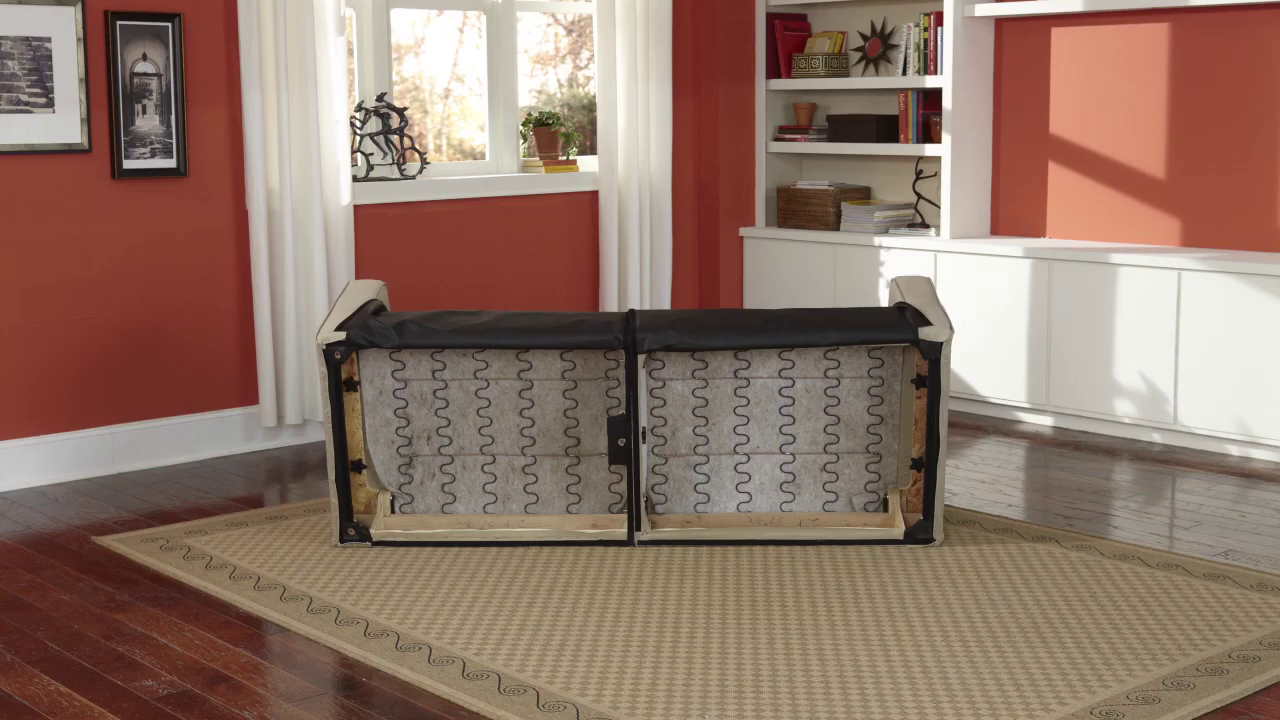}
\includegraphics[width=0.09\textwidth, height=0.1\textwidth]{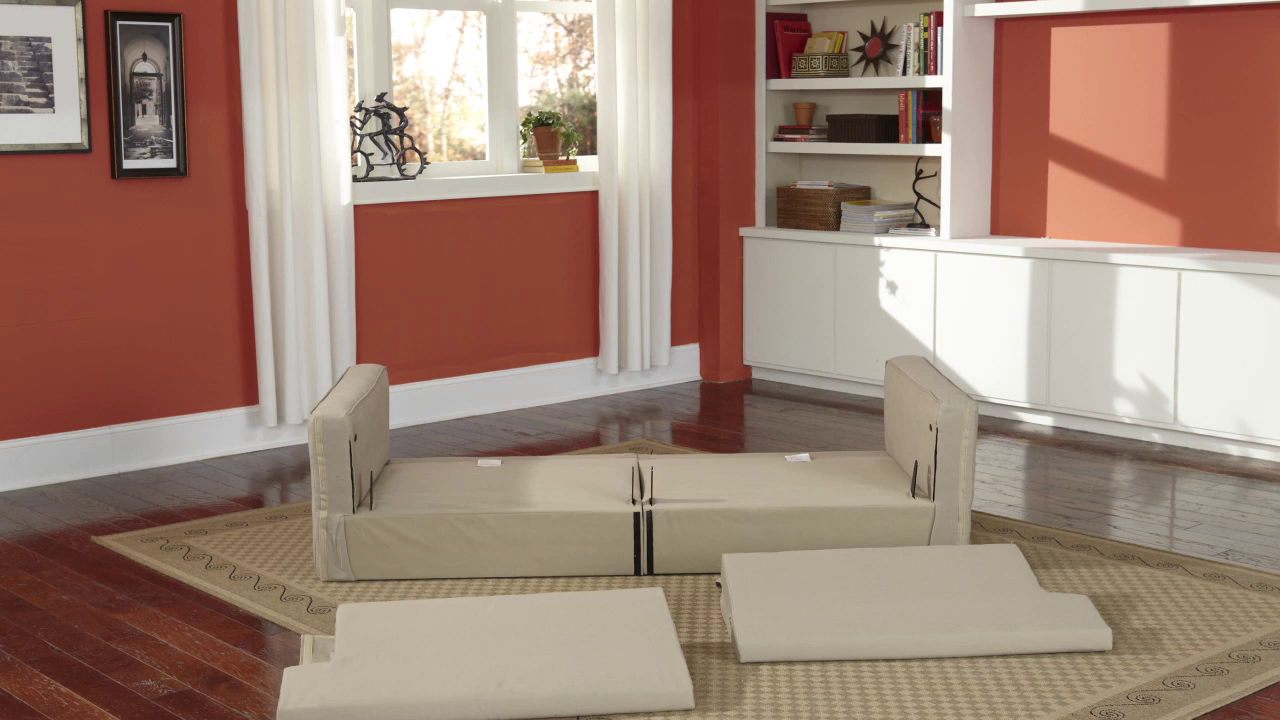}
\includegraphics[width=0.09\textwidth, height=0.1\textwidth]{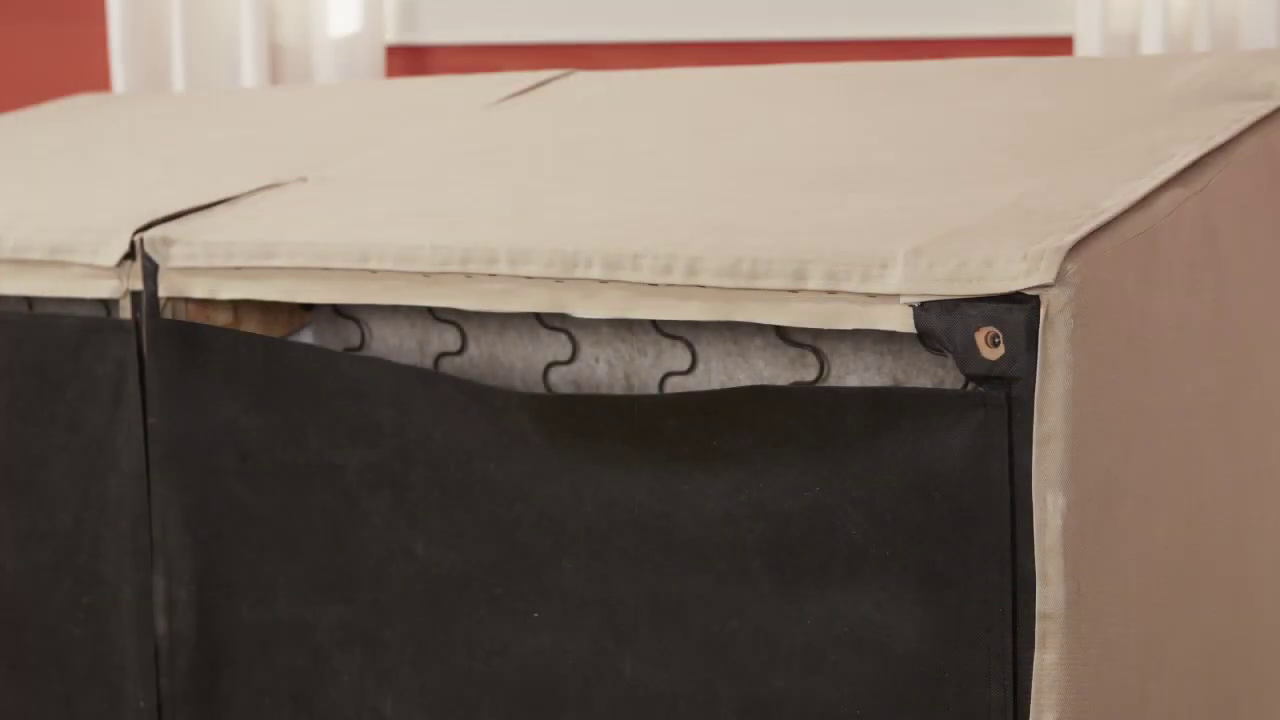}
\includegraphics[width=0.09\textwidth, height=0.1\textwidth]{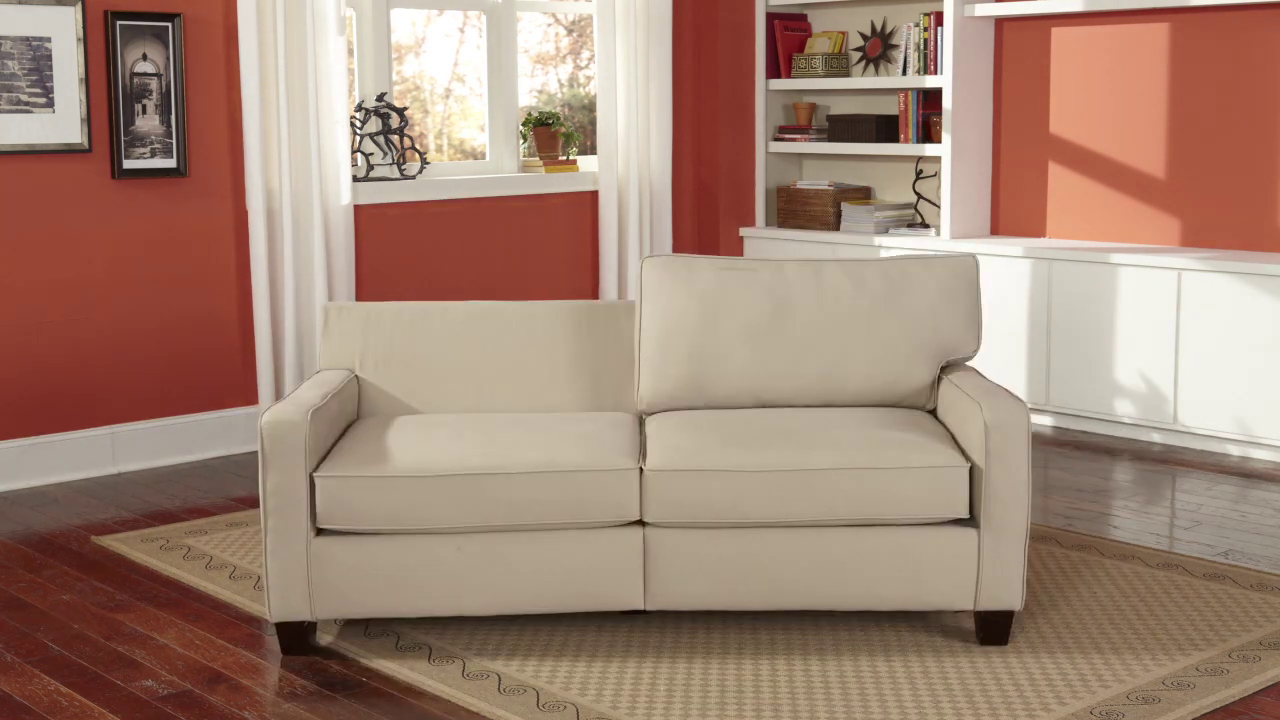}

\medskip 

\textbf{Question:} From which side is the box entering into the video, left or right?

\medskip 

\textbf{Answer:} The box is entering the video from the left side.

\caption{Example of a Spatial Understanding Question.}
\end{figure}

\begin{figure}[htbp]
\centering
\includegraphics[width=0.09\textwidth, height=0.1\textwidth]{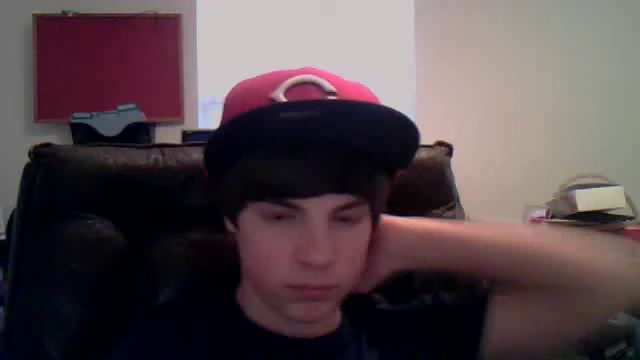}
\includegraphics[width=0.09\textwidth, height=0.1\textwidth]{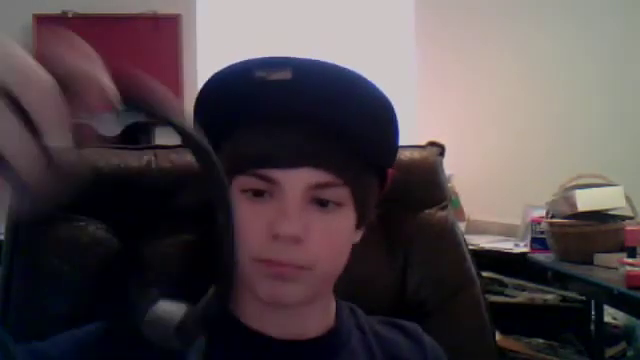}
\includegraphics[width=0.09\textwidth, height=0.1\textwidth]{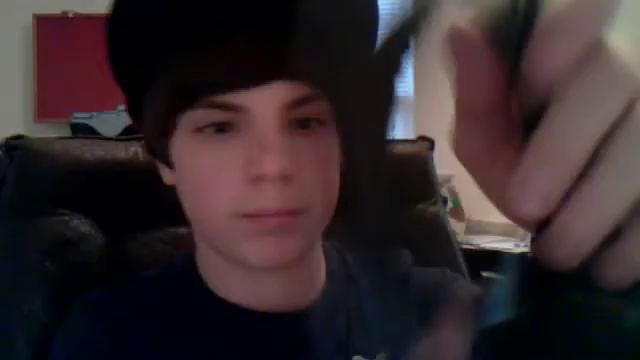}
\includegraphics[width=0.09\textwidth, height=0.1\textwidth]{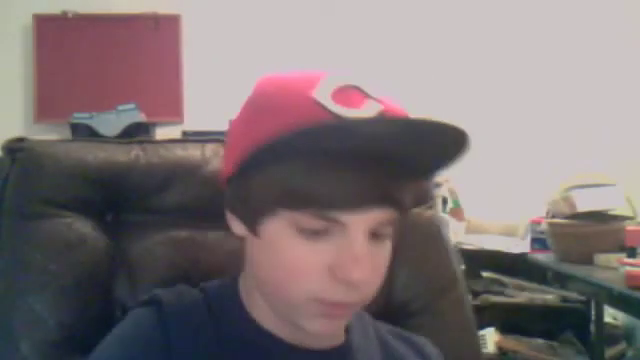}
\includegraphics[width=0.09\textwidth, height=0.1\textwidth]{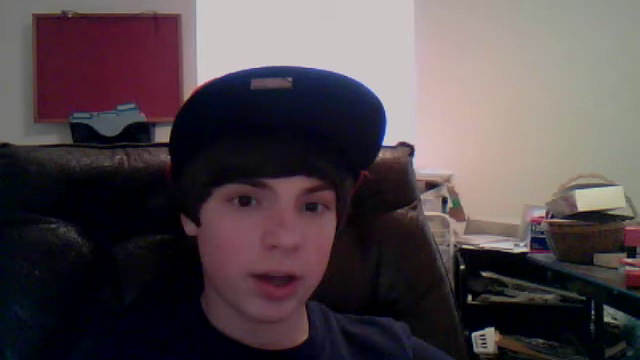}
\includegraphics[width=0.09\textwidth, height=0.1\textwidth]{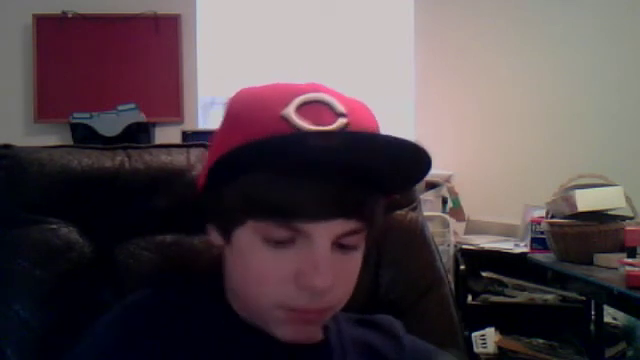}
\includegraphics[width=0.09\textwidth, height=0.1\textwidth]{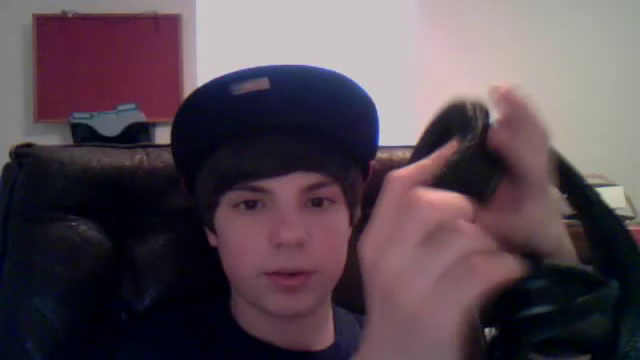}
\includegraphics[width=0.09\textwidth, height=0.1\textwidth]{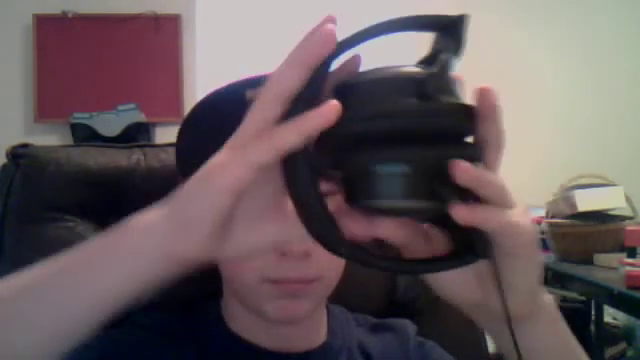}
\includegraphics[width=0.09\textwidth, height=0.1\textwidth]{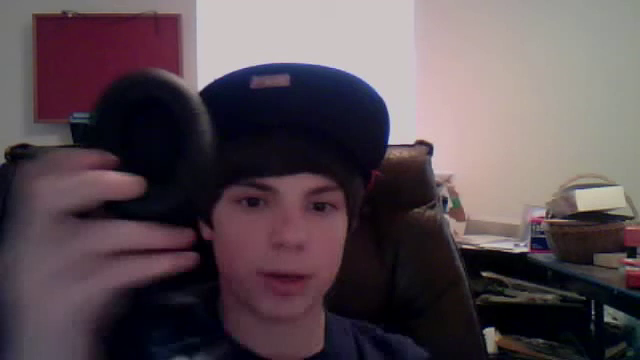}
\includegraphics[width=0.09\textwidth, height=0.1\textwidth]{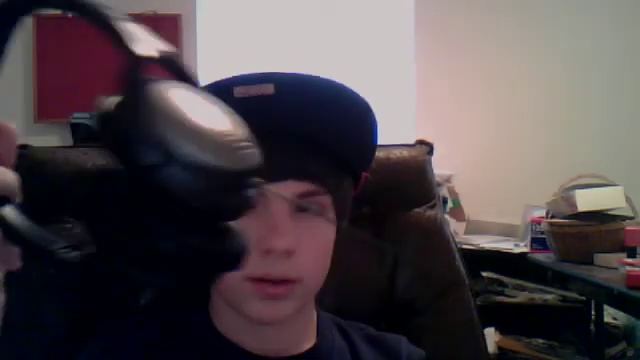}

\medskip 

\textbf{Question:} When did the boy fold the headphones to demonstrate its compactness?

\medskip 

\textbf{Answer:} He folded them at around 1 minute 47 seconds into the video.

\caption{Example of an Event and Action Recognition Question.}
\end{figure}

\begin{figure}[htbp]
\centering
\includegraphics[width=0.09\textwidth, height=0.1\textwidth]{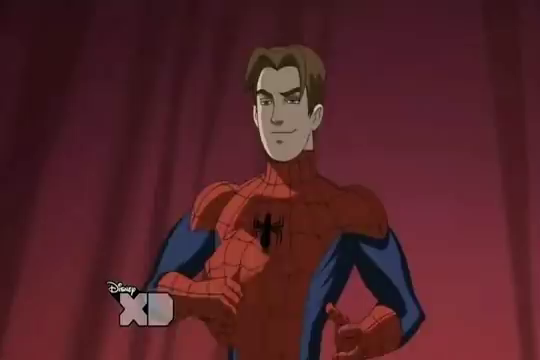}
\includegraphics[width=0.09\textwidth, height=0.1\textwidth]{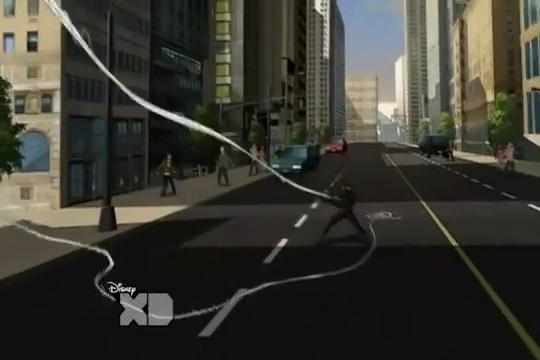}
\includegraphics[width=0.09\textwidth, height=0.1\textwidth]{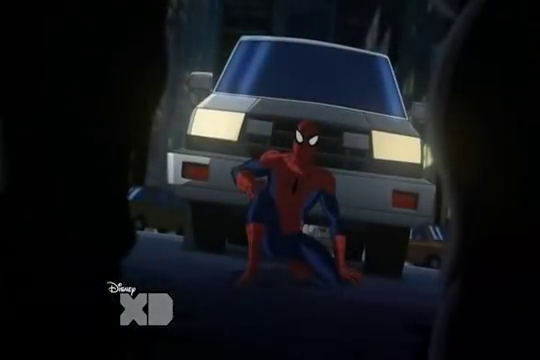}
\includegraphics[width=0.09\textwidth, height=0.1\textwidth]{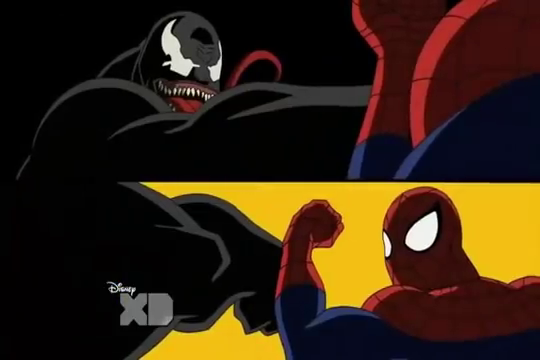}
\includegraphics[width=0.09\textwidth, height=0.1\textwidth]{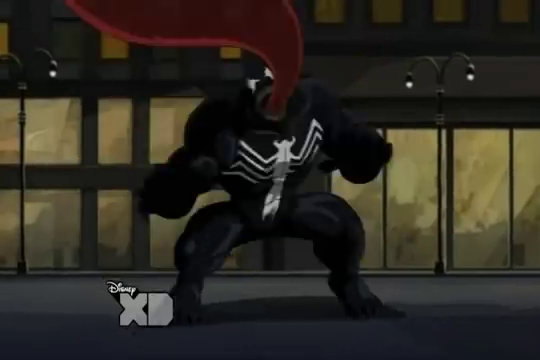}
\includegraphics[width=0.09\textwidth, height=0.1\textwidth]{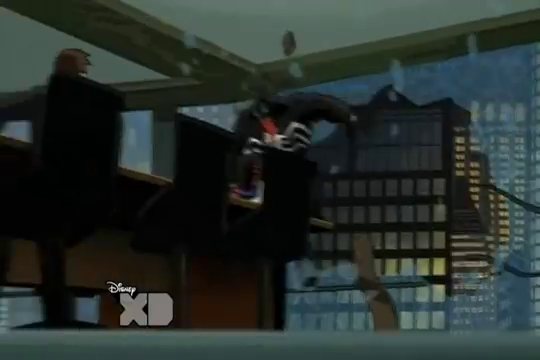}
\includegraphics[width=0.09\textwidth, height=0.1\textwidth]{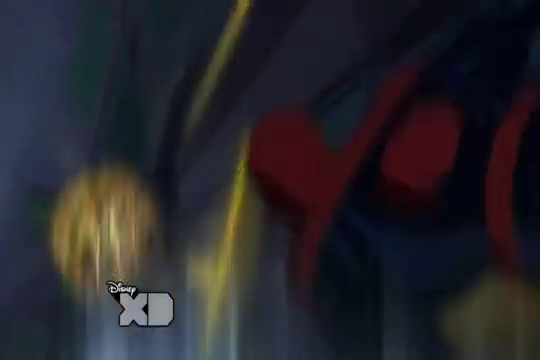}
\includegraphics[width=0.09\textwidth, height=0.1\textwidth]{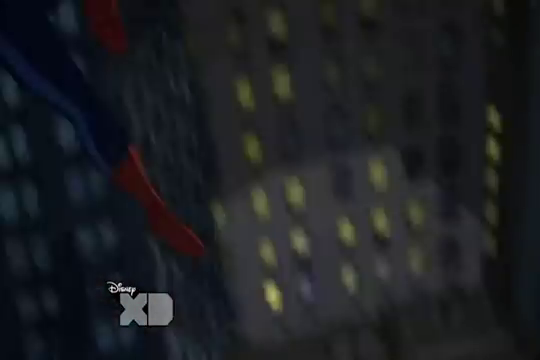}
\includegraphics[width=0.09\textwidth, height=0.1\textwidth]{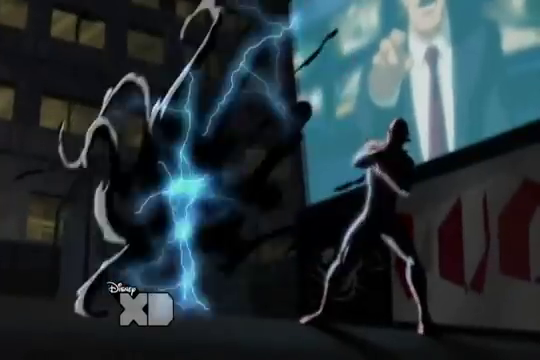}
\includegraphics[width=0.09\textwidth, height=0.1\textwidth]{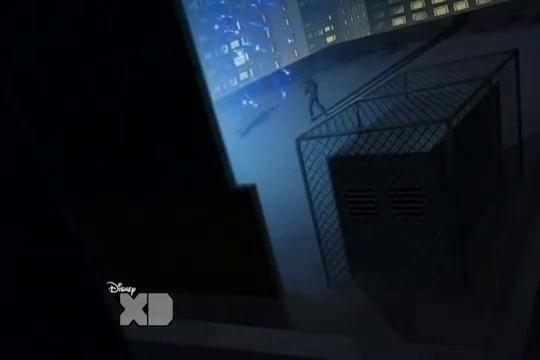}

\medskip 

\textbf{Question:} How many times was the phrase "You had to tell people" repeated throughout the video?

\medskip 

\textbf{Answer:} The phrase was repeated two times.

\caption{Example of a Dialogue and Transcript-based Question.}
\end{figure}

\begin{figure}[htbp]
\centering
\includegraphics[width=0.09\textwidth, height=0.1\textwidth]{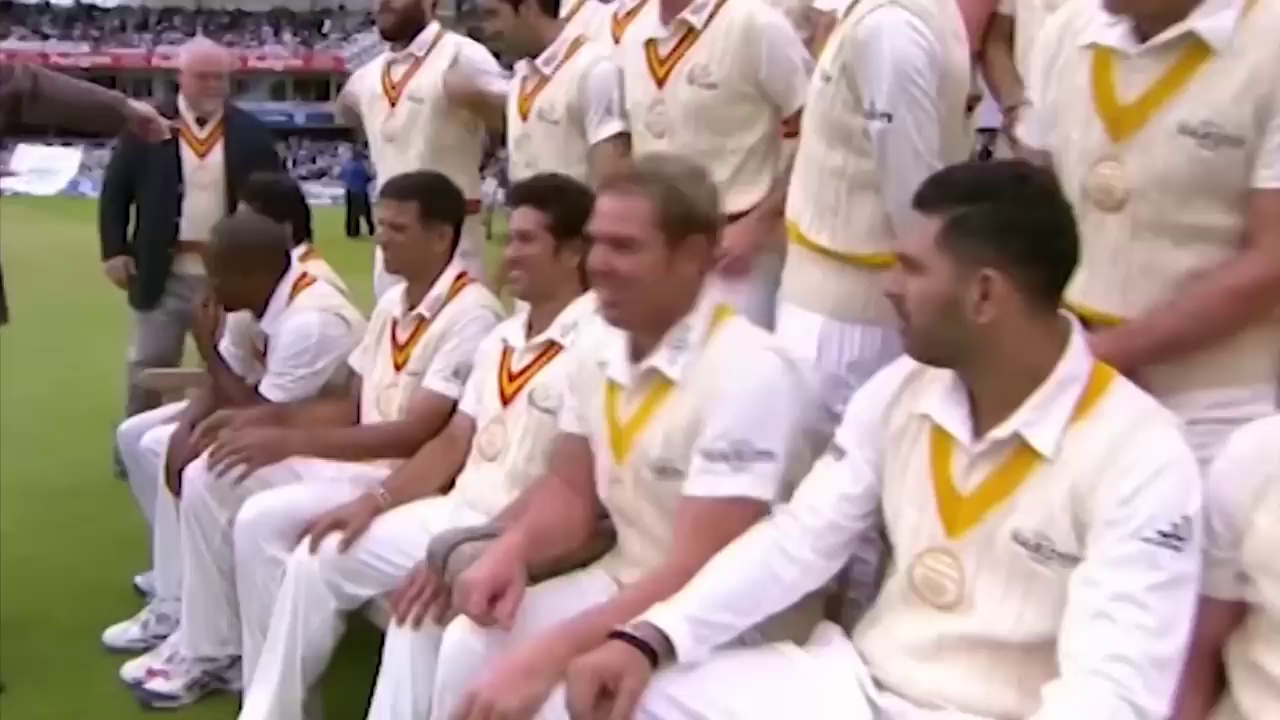}
\includegraphics[width=0.09\textwidth, height=0.1\textwidth]{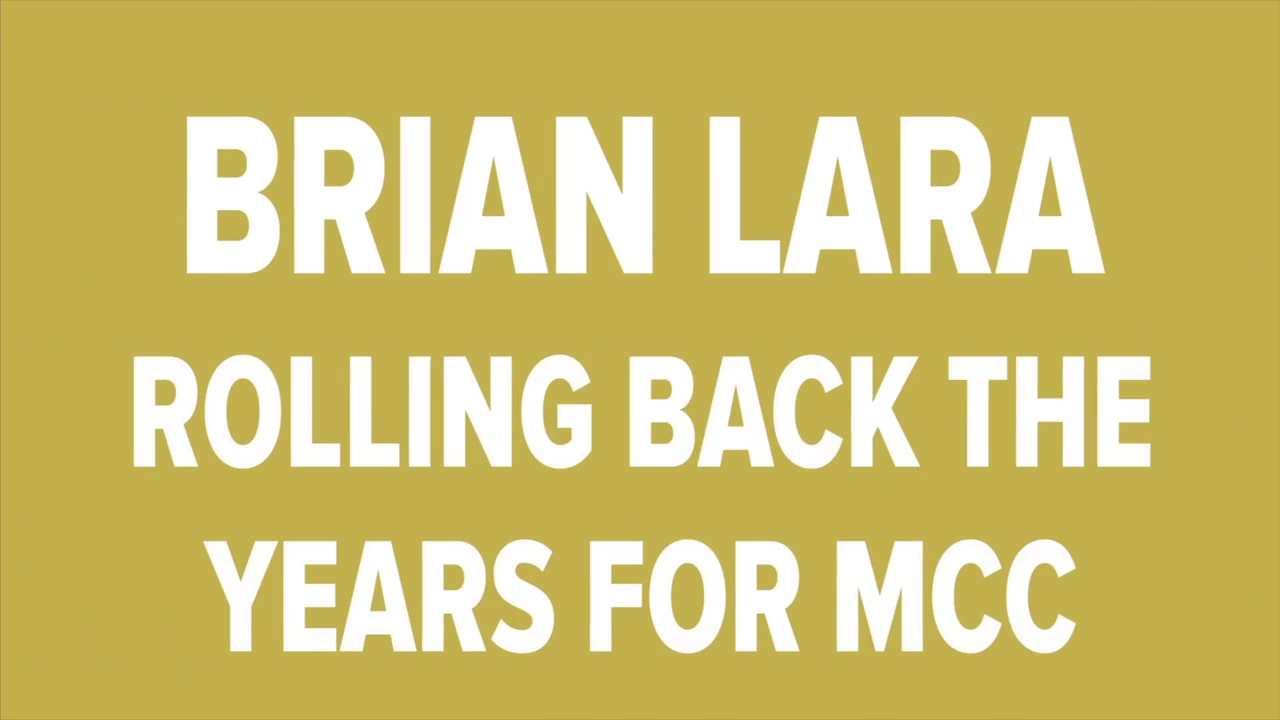}
\includegraphics[width=0.09\textwidth, height=0.1\textwidth]{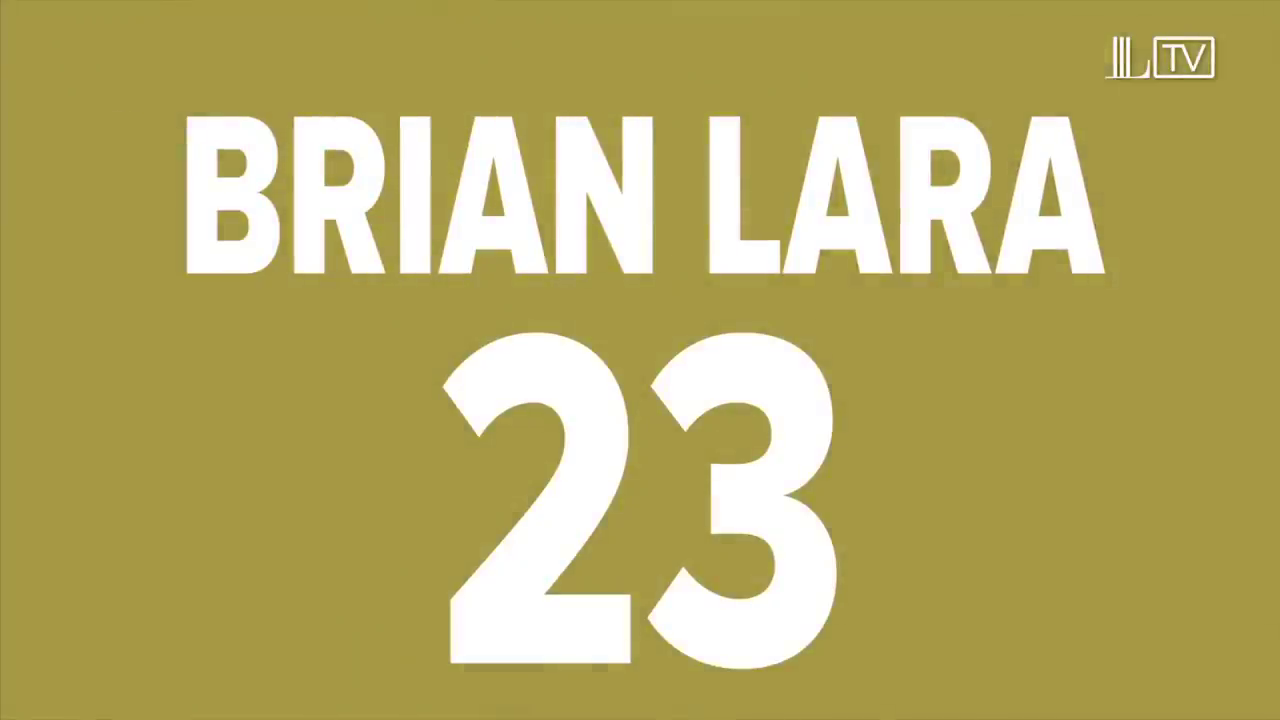}
\includegraphics[width=0.09\textwidth, height=0.1\textwidth]{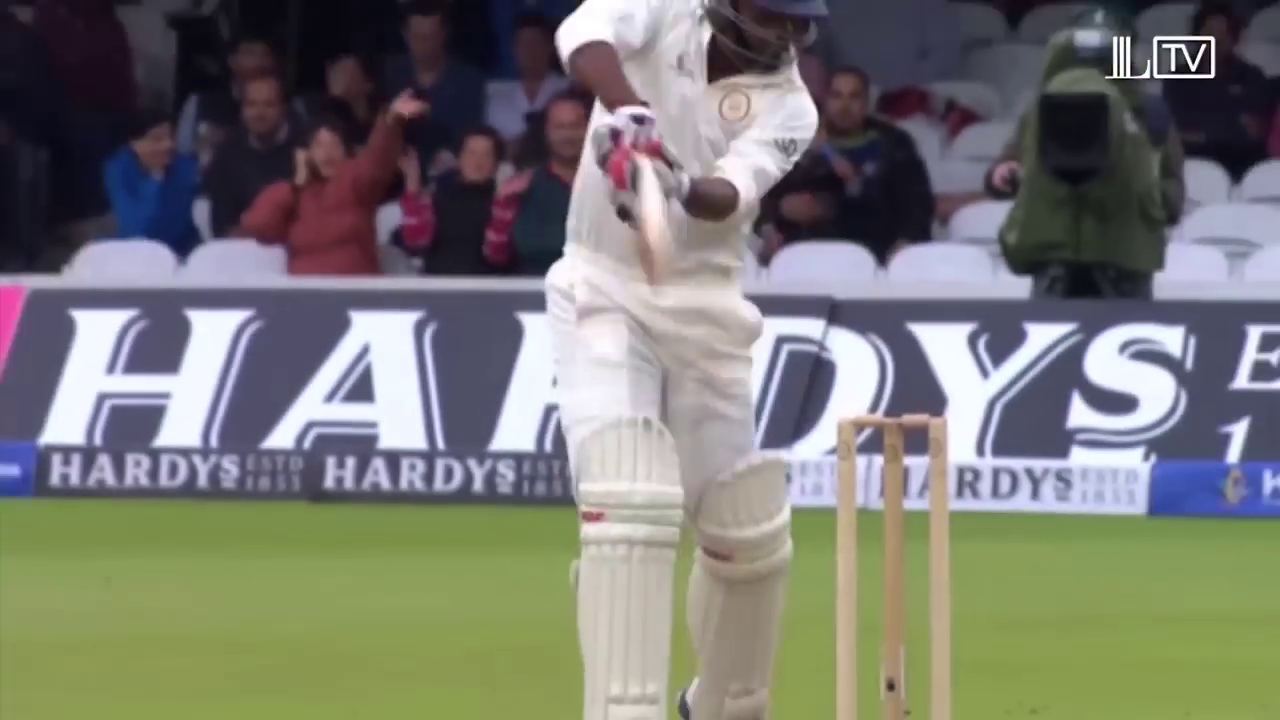}
\includegraphics[width=0.09\textwidth, height=0.1\textwidth]{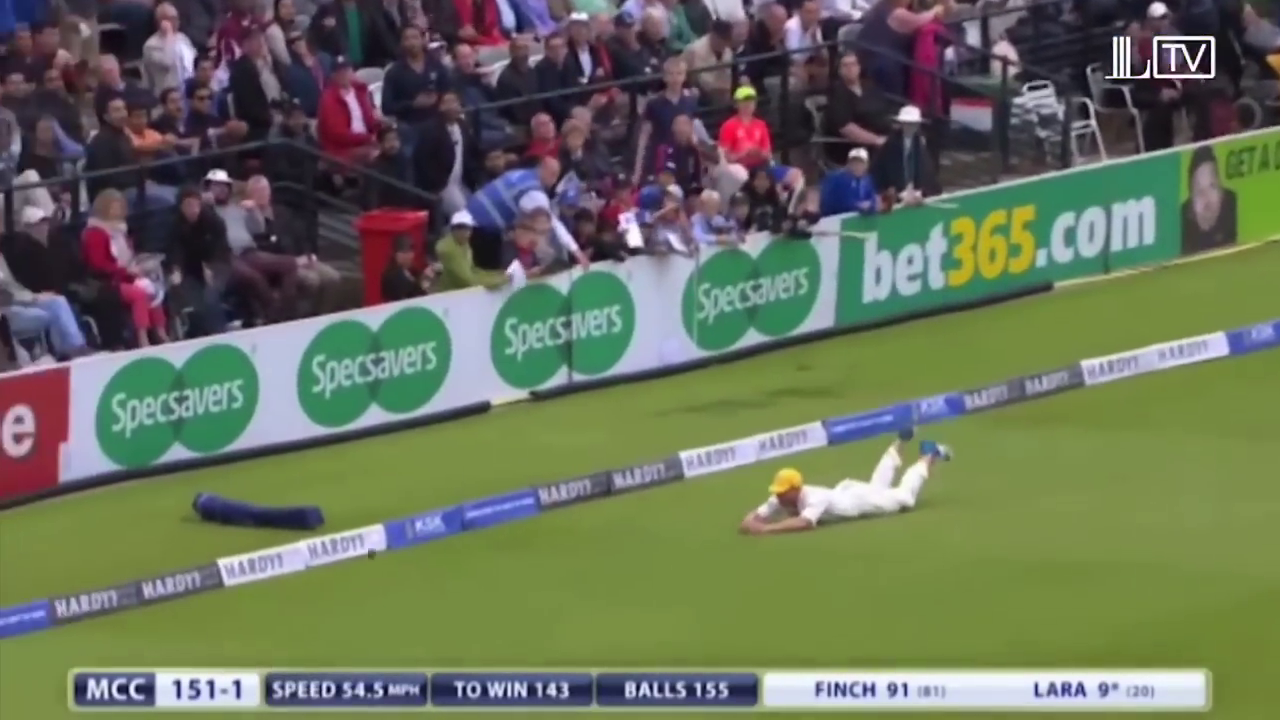}
\includegraphics[width=0.09\textwidth, height=0.1\textwidth]{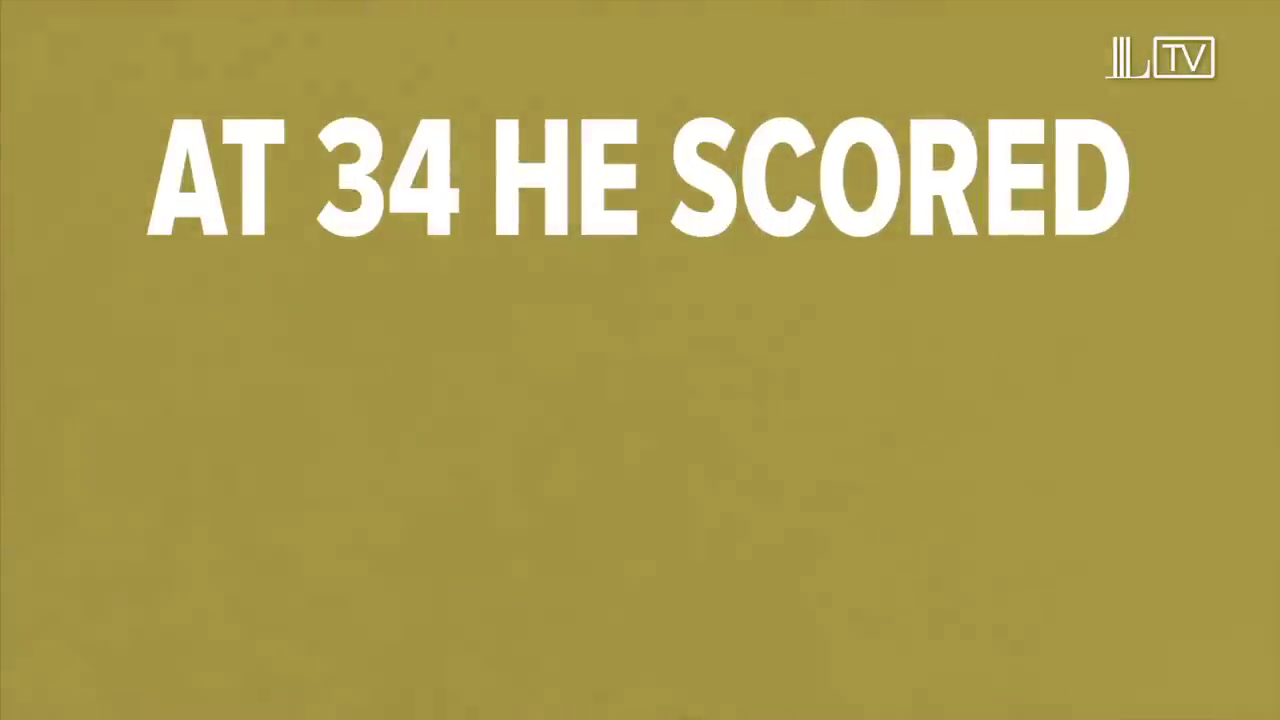}
\includegraphics[width=0.09\textwidth, height=0.1\textwidth]{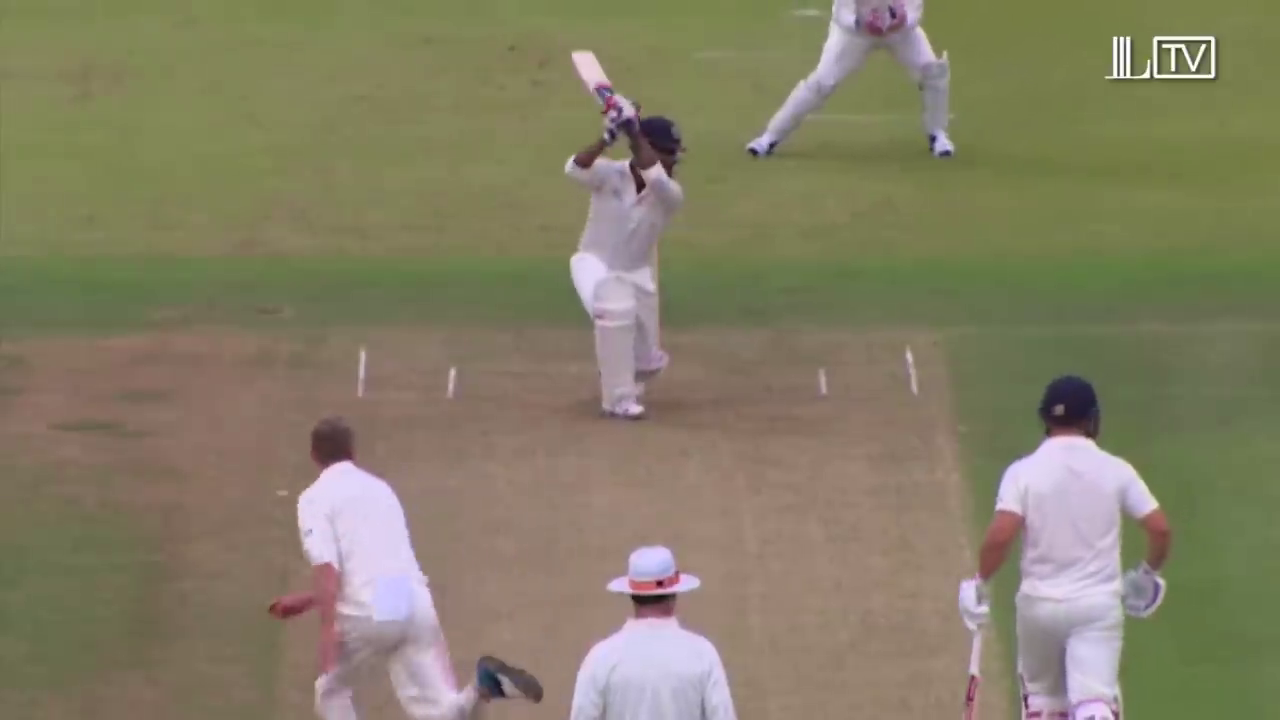}
\includegraphics[width=0.09\textwidth, height=0.1\textwidth]{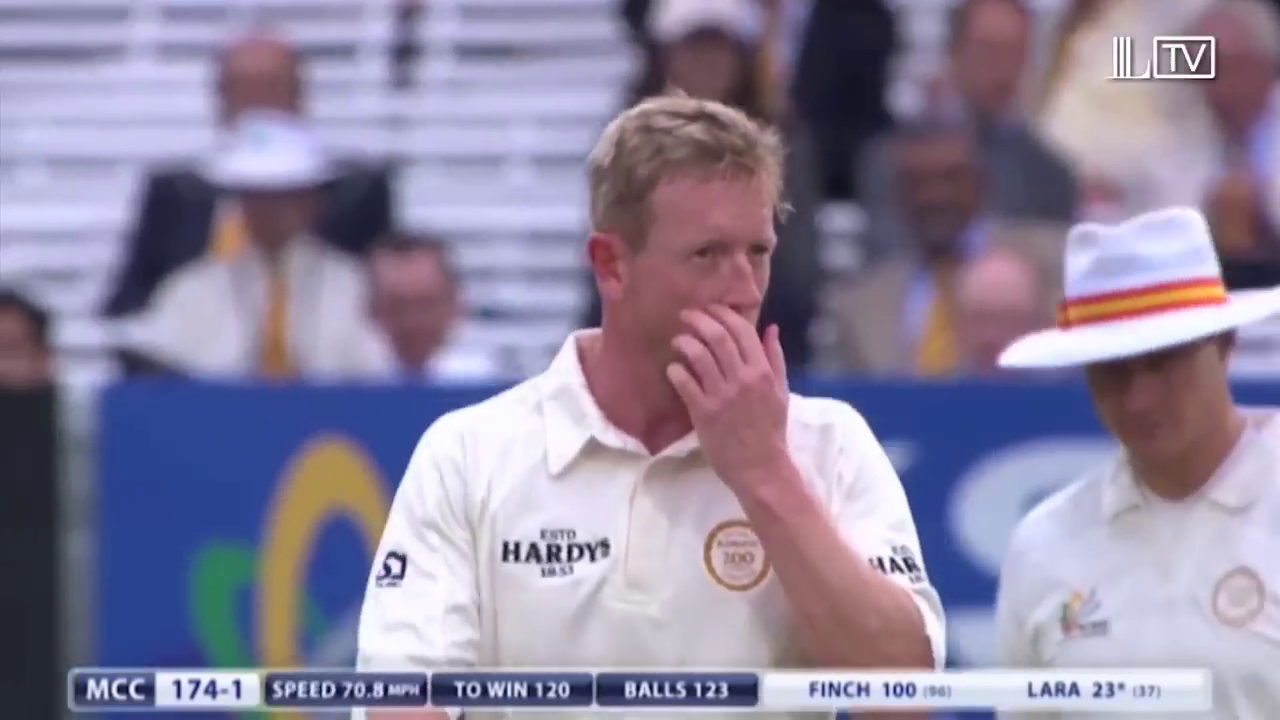}
\includegraphics[width=0.09\textwidth, height=0.1\textwidth]{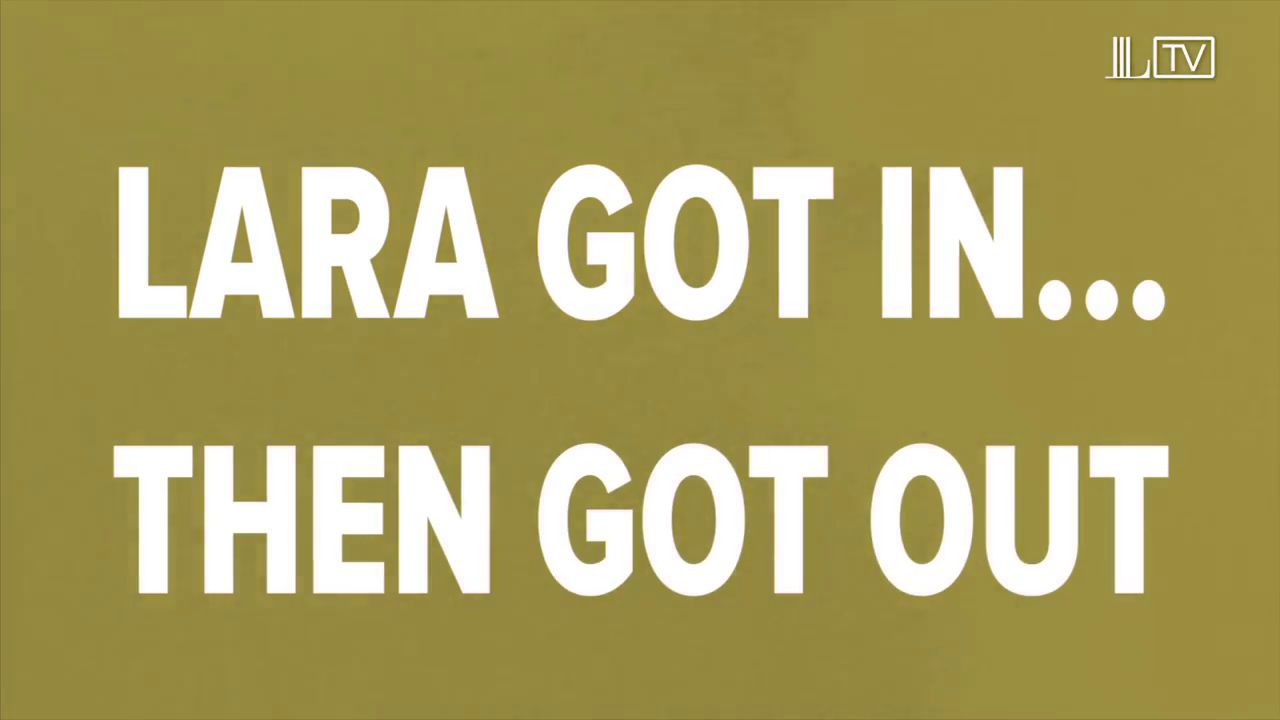}
\includegraphics[width=0.09\textwidth, height=0.1\textwidth]{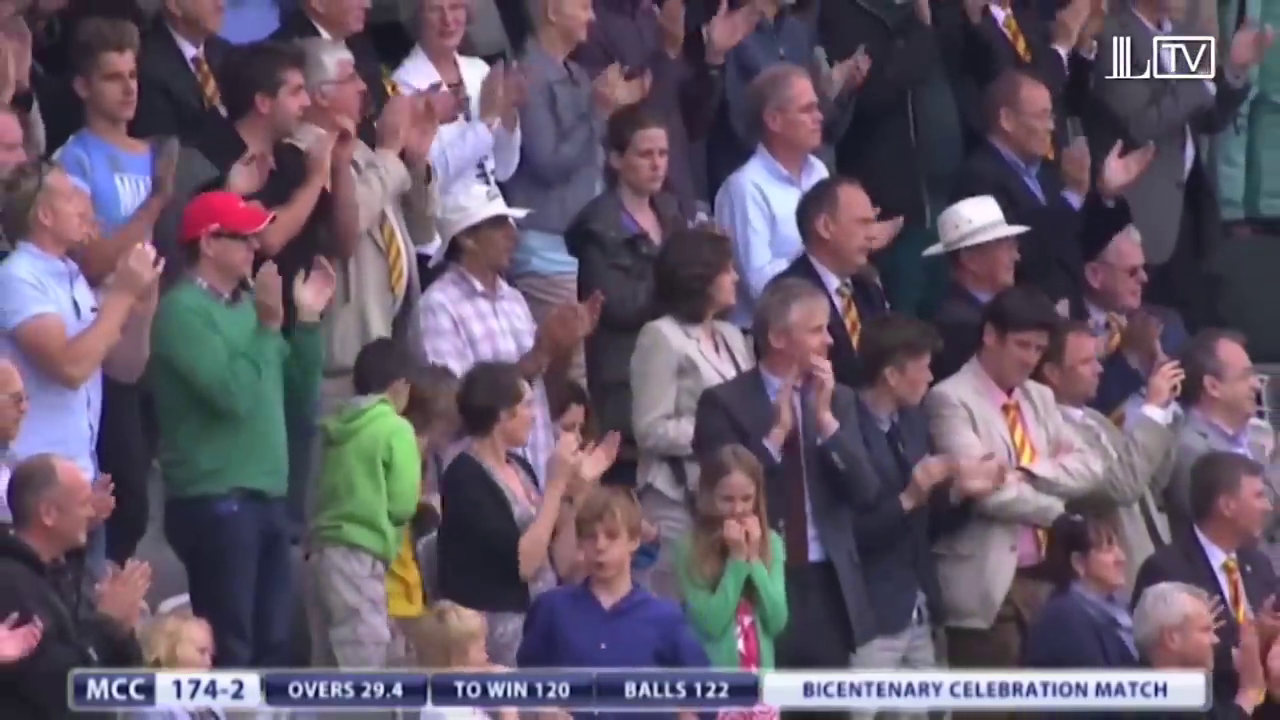}

\medskip 

\textbf{Question:} What is the vibe given off by the players in the beginning?

\medskip 

\textbf{Answer:} The players are getting ready for the match. They look calm, enthusiatic and ready.

\caption{Example of an Abstract and Conceptual Question.}
\end{figure}

\begin{figure}[htbp]
\centering
\includegraphics[width=0.09\textwidth, height=0.1\textwidth]{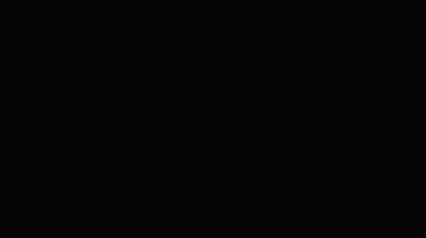}
\includegraphics[width=0.09\textwidth, height=0.1\textwidth]{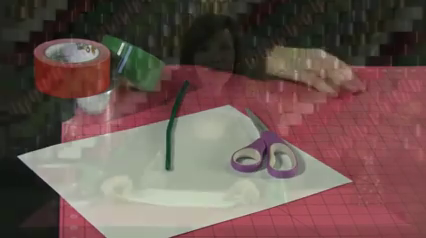}
\includegraphics[width=0.09\textwidth, height=0.1\textwidth]{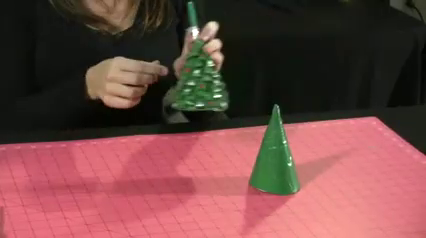}
\includegraphics[width=0.09\textwidth, height=0.1\textwidth]{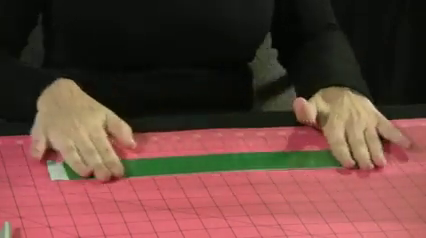}
\includegraphics[width=0.09\textwidth, height=0.1\textwidth]{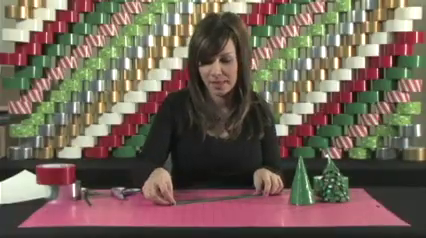}
\includegraphics[width=0.09\textwidth, height=0.1\textwidth]{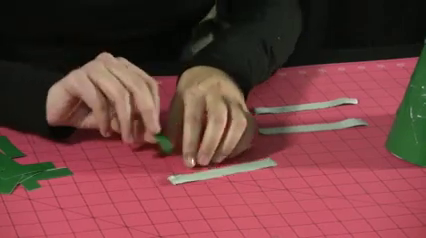}
\includegraphics[width=0.09\textwidth, height=0.1\textwidth]{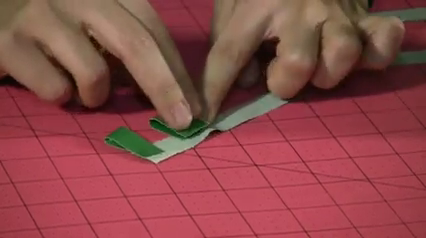}
\includegraphics[width=0.09\textwidth, height=0.1\textwidth]{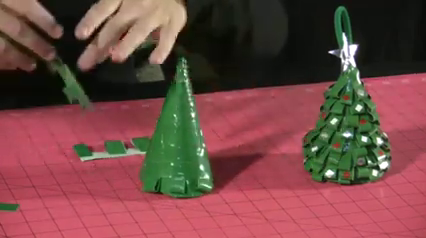}
\includegraphics[width=0.09\textwidth, height=0.1\textwidth]{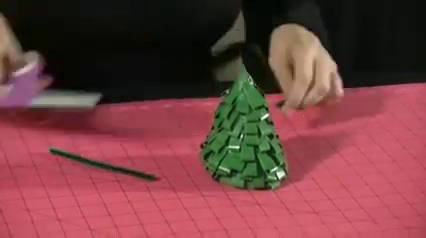}
\includegraphics[width=0.09\textwidth, height=0.1\textwidth]{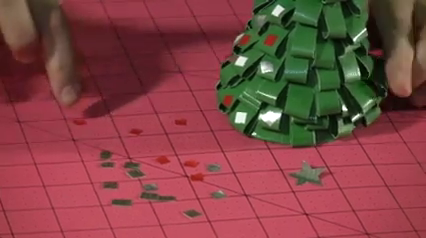}

\medskip 

\textbf{Question:} What is the colour of the scissors that is on the table?

\medskip 

\textbf{Answer:} The colour of the scissors on the table is purple.

\caption{Example of a Specific Detail Based Question.}
\end{figure}
\subsection*{Evaluation in MMCTQA}
We use GPT4-Turbo with the following prompt to evaluate the performance of individual samples in MMCTQA:
\begin{lstlisting}
You are an evaluator for a video question answering system. You will be given the following things:
<given>
Question: A question on a video.
Ground Truth Answer: Answer annotated by a human.
System Answer: Answer from System.
</given>

Your job is to label the System Answer as Correct, Incorrect, or Partially Correct.
To effectively assess the system answer, use the following criteria to determine whether the answer is "Correct", "Incorrect", or "Partially Correct":
<criteria>
Correct: The answer should fully capture the main theme or essential details of the ground truth answer. For factual questions, this means including all critical facts, but minor details can be omitted without affecting the verdict. For questions asking for a specific moment or timestamp, a 5-second leeway between the ground truth and the answer is acceptable. For descriptive questions, the response should accurately reflect the essence and details of the ground truth, and may include additional relevant explanations that align with the theme of the ground truth. If the response is mostly accurate and any missing elements do not significantly change the understanding, it should be considered Correct.
Partially Correct: The answer captures significant aspects of the ground truth but misses one or more critical components or details that alter the fundamental understanding or facts of the response.
Incorrect: The answer fails to correctly address the ground truth. This could be due to major factual errors, significant incomplete information, or a fundamental misunderstanding of the main theme or key details.
</criteria>
Evaluate the system answer based on these guidelines to determine its accuracy and completeness in relation to the ground truth provided for each question. You must respond as follows:
<response_format>
System Answer: [Verdict]
</response_format>
Here [Verdict] can be one of "Correct", "Incorrect", or "Partially Correct". You should only respond in this format with one line and the verdict as one of the given options. No extra lines and no extra text whatsoever.
\end{lstlisting}
\section{Vision-based Critic Performance}
\label{app:critic_quali}

\begin{figure*}[t!]
\begin{minipage}[t]{0.49\linewidth}
        \includegraphics[trim=0.7cm 1cm 1cm 1cm, clip, width=0.95\columnwidth]{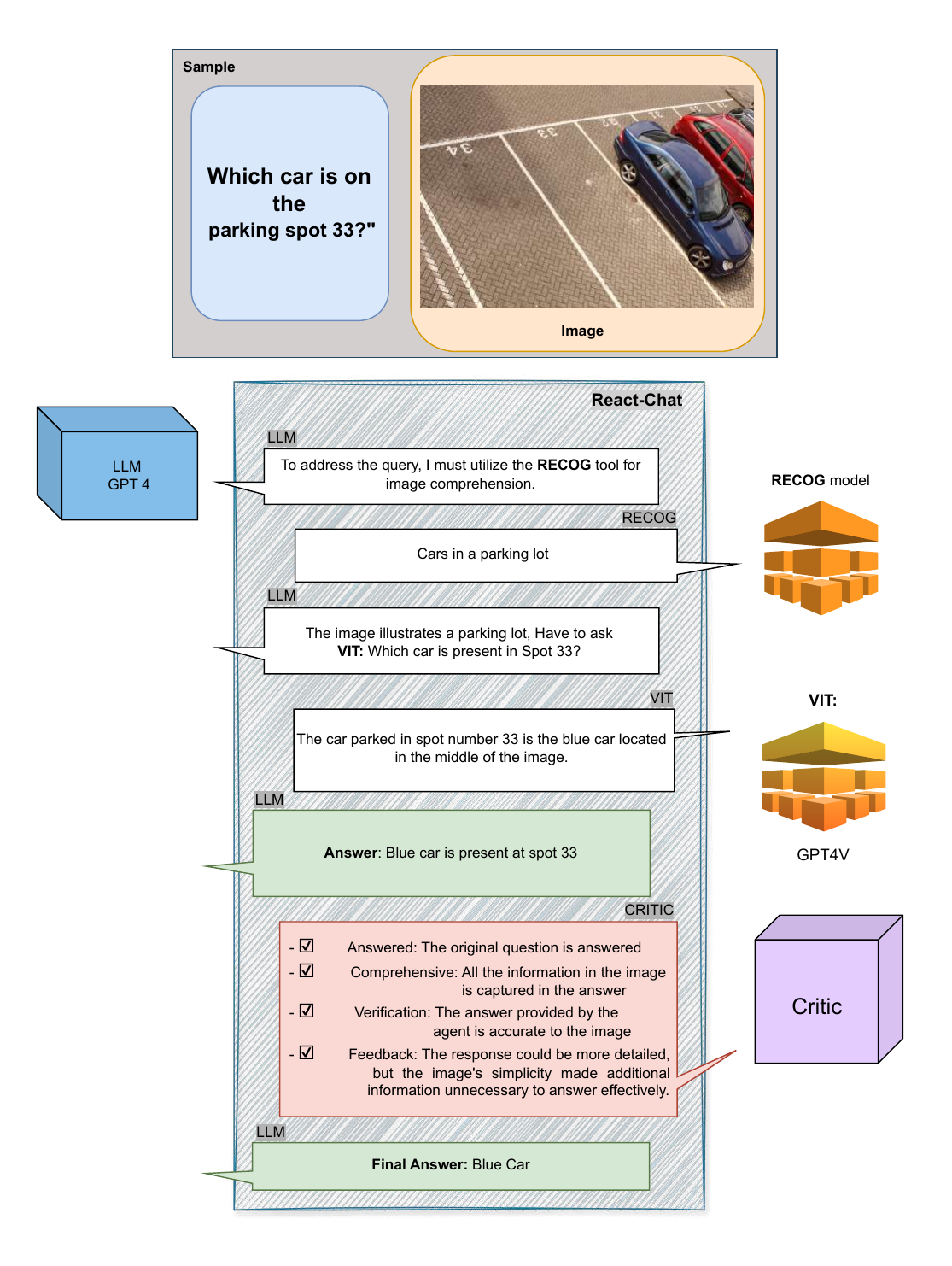}
        \caption{A reasoning chain demonstrating Failure Case I }
        \label{fig:base_fault_img_example}
        \vspace{-5pt}
\end{minipage}
\hspace{5pt}
\begin{minipage}[t]{0.49\linewidth}
        \includegraphics[trim=0.7cm 1.1cm 1cm 1cm, clip, width=0.95\columnwidth ]{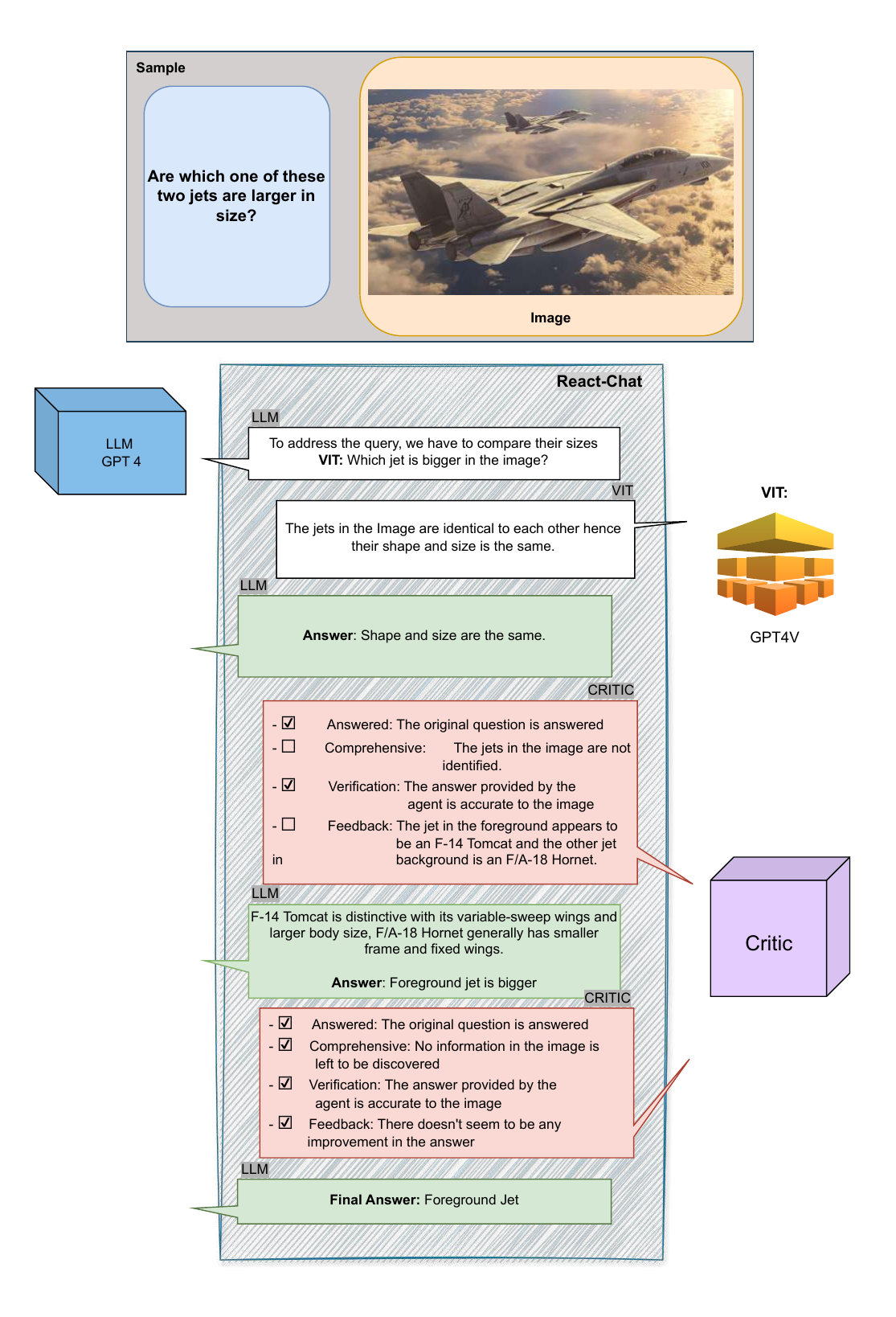}
        \caption{A reasoning chain demonstrating Failure Case II}
        \label{fig:critic_fault_img_example}
        \vspace{-5pt}
\end{minipage}

\end{figure*}

As previously learnt there are multiple scenarios where our Pipeline is at fault. Critic Improves these critical points in the pipeline but doesn’t completely mitigate them. We have scene how Critic Fixes VIT’s output in Figure. \ref{fig:image_example}, there are still 2 more cases where such errors are caused 1) When the base pipeline proposes a wrong answer and Critic accepts the wrong answer that is 20.41\% of the samples. 2) When the pipeline would have reached the right answer but the critic made it choose the wrong answer this has happened 5.35\% of the samples. Hence it is very critical to understand individual of these critical points to make a note of for future work.

\subsection*{Base Pipeline wrong and Critic Wrong}
	This majorly due to the underperformance of the Vision Language Model we choose, as both the models are GPT-4-Vision for VIT and Critic they share common weak point especially in celebrity detection, OCR and hallucinations. We present you few examples that demonstrate these weaknesses of our pipeline. In Figure. \ref{fig:base_fault_img_example} we see a simple question utilizing Spatial understanding and OCR capabilities, as seen in the image car spot is empty and the answer was “No<OR>empty”. But as seen in the chat VIT model misinterprets the number over the blue car and the Critic also identifies the same and doesn’t give useful feedback. This could be because of the inverted numbers where difficult for the model to read or understand the trend in the number of the spot making it believe its answer or the camera angle is skewed which made the Vision Language model to hallucinate the answer. But it is certain in cases like these where the smallest sub unit of question is difficult to answer by the Vision Language model it fails to recognize to critic it or use any other tool.

\subsection*{Critic Leads the base pipeline to the wrong answer}
	There are very few samples where the introduction of the critic leads to the wrong answer, These samples are very interesting as they give better insight into LLM hallucination due to intent. We can see the example Figure. \ref{fig:critic_fault_img_example} where the problem is to identify the bigger jet in the image, the image contains identical jets flying together and which was correctly identified by the base pipeline but Critic tries hard to differentiate between the jet and infers the jets as F-14 Tomcat and F/a-18 super hornet which are very similar to each other in shape and appearance except for the size. This could be a good quality of the pipeline or even a bad behavior where it doesn’t choose simplicity over specifics. Other samples under this category are due to hallucination of the base pipeline for being familiar with the question type and image, dismissing to evaluate individual details causing the Critic to control the pipeline’s output.


\end{document}